\documentclass[letterpaper, 10 pt, journal, twoside]{IEEEtran}   

\usepackage{times}
\usepackage{epsfig}
\usepackage{graphicx}
\usepackage{amsmath,amsfonts,amssymb,amsthm}
\usepackage{color}
\usepackage{caption}
\usepackage{subcaption}
\usepackage{xspace}
\usepackage{array}
\usepackage{cite}
\usepackage{float}
\usepackage{multirow}
\usepackage[pagebackref=true,breaklinks=true,letterpaper=true,colorlinks,bookmarks=false]{hyperref}
\usepackage[utf8]{inputenc}
\usepackage{multirow}
\usepackage[normalem]{ulem}
\usepackage{booktabs}
\usepackage{multibib}

\newcommand{\bB}{\mathbf{B}}

\newcommand{\bI}{\mathbf{I}}

\newcommand{\bu}{\mathbf{u}}\newcommand{\bU}{\mathbf{U}}

\newcommand{\bx}{\mathbf{x}}

\newcommand{\nR}{\mathbb{R}}

\newcommand{\cL}{\mathcal{L}}

\newcommand{\cW}{\mathcal{W}}

\newcommand{\figref}[1]{Fig.~\ref{#1}}
\newcommand{\secref}[1]{Section~\ref{#1}}

\newcommand{\tabref}[1]{Table~\ref{#1}}

\makeatletter
\DeclareRobustCommand\onedot{\futurelet\@let@token\@onedot}
\def\@onedot{\ifx\@let@token.\else.\null\fi\xspace}
\def\eg{e.g\onedot} 
\def\ie{i.e\onedot}

\def\etal{et~al\onedot}

\makeatother

\newcommand{\boldparagraph}[1]{\vspace{0.2cm}\noindent{\bf #1:} }
\newcommand{\PAR}[1]{\vspace{0.2cm}\noindent{\bf #1} }

\definecolor{darkgreen}{rgb}{0,0.7,0}

\newcommand{\stkout}[1]{\ifmmode\text{\sout{\ensuremath{#1}}}\else\sout{#1}\fi}

\begin{document}
\renewcommand{\thefootnote}{\alph{footnote}}

\title{Self-Supervised Linear Motion Deblurring}

\author{Peidong Liu$^{1}$, Joel Janai$^{2}$, Marc Pollefeys$^{1,3}$, Torsten Sattler$^{4}$ and Andreas Geiger$^{2}$%
	\thanks{Manuscript received: September, 10, 2019; Revised December, 06, 2019; Accepted January, 20, 2020.}
	\thanks{This paper was recommended for publication by Editor Cesar Cadena Lerma upon evaluation of the Associate Editor and Reviewers’ comments.} 
	\thanks{$^{1}$ Peidong Liu and Marc Pollefeys are with Computer Vision and Geometry Group, Department of Computer Science, ETH Z\"{u}rich, Switzerland {\tt\footnotesize peidong.liu@inf.ethz.ch}}%
	\thanks{$^{2}$ Joel Janai and Andreas Geiger are with Autonomous Vision Group, Max Planck Institute for Intelligent Systems and Univeristy of T\"{u}bingen, T\"{u}bingen, Germany}%
	\thanks{$^{3}$ Marc Pollefeys is with Microsoft Mixed Reality and Artificial Intelligence Lab, Z\"{u}rich, Switzerland}
	\thanks{$^{4}$ Torsten Sattler is with Computer Vision and Medical Image Analysis Group, Chalmers University of Technology, Sweden}
	\thanks{Digital Object Identifier (DOI): see top of this page.}
}

\maketitle


\begin{abstract}
Motion blurry images challenge many computer vision algorithms, \eg, feature detection, motion estimation, or object recognition. Deep convolutional neural networks are state-of-the-art for image deblurring. However, obtaining training data with corresponding sharp and blurry image pairs can be difficult. In this paper, we present a differentiable reblur model for self-supervised motion deblurring, which enables the network to learn from real-world blurry image sequences without relying on sharp images for supervision. Our key insight is that motion cues obtained from consecutive images yield sufficient information to inform the deblurring task. We therefore formulate deblurring as an inverse rendering problem, taking into account the physical image formation process: we first predict two deblurred images from which we estimate the corresponding optical flow. Using these predictions, we re-render the blurred images and minimize the difference with respect to the original blurry inputs. We use both synthetic and real dataset for experimental evaluations. Our experiments demonstrate that self-supervised single image deblurring is really feasible and leads to visually compelling results. Both the code and datasets are available at \href{https://github.com/ethliup/SelfDeblur}{https://github.com/ethliup/SelfDeblur}. 
\end{abstract}



\section{Introduction}
\label{sec:intro}

\begin{figure*}[t!]
\centering
\minipage{0.25\textwidth}
\includegraphics[width=\linewidth]{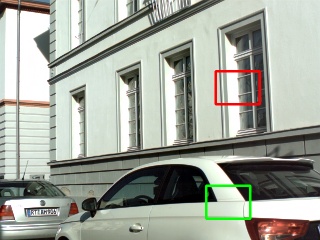}
\endminipage\hfill
\minipage{0.25\textwidth}
\includegraphics[width=\linewidth]{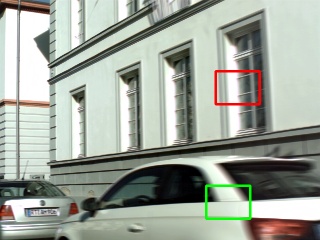}
\endminipage\hfill
\minipage{0.25\textwidth}
\includegraphics[width=\linewidth]{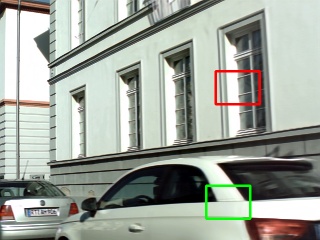}
\endminipage\hfill
\minipage{0.25\textwidth}
\includegraphics[width=\linewidth]{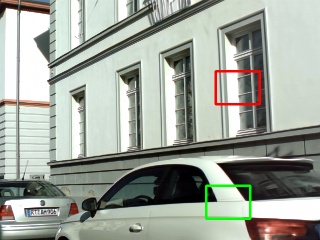}
\endminipage\vfill
\minipage{0.25\textwidth}
\includegraphics[width=\linewidth]{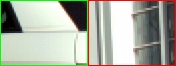}
\endminipage\hfill
\minipage{0.25\textwidth}
\includegraphics[width=\linewidth]{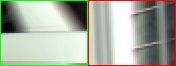}
\endminipage\hfill
\minipage{0.25\textwidth}
\includegraphics[width=\linewidth]{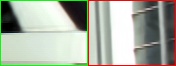}
\endminipage\hfill
\minipage{0.25\textwidth}
\includegraphics[width=\linewidth]{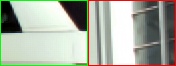}
\endminipage
\caption{\textbf{Self-supervised motion deblurring.} \textbf{First:} Sharp ground truth image. \textbf{Second:} Blurry input image. \textbf{Third:} Deblurring results of our self-supervised method. \textbf{Fourth:} Deblurring results of the supervised method from Tao et al. \cite{Tao2018CVPR}.}
\label{fig_teaser_image}
\end{figure*}

\IEEEPARstart{M}{otion} blur is one of the most common factors degrading image quality.
It often arises when the image content changes quickly (\eg, due to fast camera motion) or when the environment is illuminated poorly, hence necessitating longer exposure times. Combining both situations, \eg, a self-driving car driving at dusk, further aggravates the problem. As many computer vision algorithms such as semantic segmentation, object detection, or visual odometry rely on visual input, blurry images challenge the performance of these algorithms.
It is well known that many algorithms (\eg, depth prediction, feature detection, motion estimation, or object recognition) suffer from motion blur \cite{Pretto2009ICRA, Kupyn2018CVPR, Tourani2016ICRA, Qiu2019CVPR}. The motion deblurring problem has thus received considerable attention in the past~\cite{Shan2008ToG,Gupta2010ECCV,Nah2017CVPR,Tao2018CVPR,Kupyn2018CVPR}.

Existing techniques to solve this problem can be classified into two categories: 
the first type of approaches formulate the problem as an optimization problem \cite{Krishnan2009NIPS,Xu2010ECCV, Fergus2006SIGGGRAPH, Shan2008ToG, Levin2009CVPR, Cho2009ToG} where the latent sharp image and/or the blur kernel are optimized using gradient descent. One of the advantages of this kind of methods is that they do not require any ground truth sharp images. However, the resulting solvers usually have a large computational complexity, which limits their applicability in time-constrained settings, such as real-time robotic visual perception. Handcrafted priors on either the image or the blur kernel further limit their performances. 

The second type of approaches phrase the task as a learning problem. Building upon the recent advances of deep convolutional neural networks, state-of-the-art results have been obtained for both single image deblurring \cite{Nah2017CVPR, Tao2018CVPR} and video deblurring \cite{Su2017CVPR}, outperforming optimization-based techniques in terms of both quality and efficiency.
However, learning-based methods typically require full supervision in the form of corresponding pairs of blurred and sharp images.  
Unfortunately, obtaining such pairs is not always easy due to two main reasons. One is that not every camera has the capability to capture images at enough high frame rate (1000 or more frames per second), such that we can use the recorded frames to synthesize the training data. Another reason is that it would also be difficult to obtain good quality images in real scenarios where the motion blur really occurs (\eg, at night). High frame rate limits the exposure time and would thus make the captured image extremely dark or even invisible.

Inspired by recent progress in self-supervised depth \cite{Zhang2018ECCV,Godard2017CVPR}, flow \cite{Meister2018AAAI,Janai2018ECCV} and representation learning \cite{Pathak2016CVPR,Doersch2015ICCV}, we propose a novel approach for self-supervised image deblurring which only relies on real-world blurry image sequences for training. Self-supervised learning improves the network's generalization performance, by enabling the network to adapt to scenarios where ground truth sharp images are not available.
Our network contains a deblurring network and an optical flow estimation network.
However, instead of using ground truth sharp images\cite{Nah2017CVPR, Tao2018CVPR, Kupyn2018CVPR, Ilg2017CVPR}, we pose the task as an inverse rendering problem and take advantage of the physical image formation process for supervision. More specifically, given two consecutive blurry frames of a video sequence, we first predict the corresponding deblurred images using a deep neural network. A second deep neural network takes both deblurred images as input and computes the corresponding optical flow. Using this prediction, and assuming a locally linear blur kernel, our model re-renders the blurred images and compares the results to the original blurry inputs using a photometric loss function. Moreover, we constrain the optical flow network using a photo-consistency loss function. Our entire model can be trained end-to-end from pairs of consecutive blurry images captured with a consumer video camera. At test time, our network takes a single blurred image and deblurs it in real time on a single GTX 1080Ti graphic card using the learned parameters. As illustrated in \figref{fig_teaser_image}, our approach is competitive with respect to a state-of-the-art supervised method \cite{Tao2018CVPR} despite being fully self-supervised. 

Our second contribution is a novel synthetic dataset and a real dataset. The synthetic dataset has 3606 blurry-sharp image pairs recorded with a professional high-speed camera mounted on a ground vehicle. The real dataset has 2302 blurry images and is recorded with a normal camera. We use both datasets to evaluate our algorithm against several baselines both quantitatively and qualitatively. 

\section{Related Work}
\label{sec:related}
Motion deblurring methods can be categorized into two groups: those that assume spatially uniform blur \cite{Richardson1972,Krishnan2009NIPS,Xu2010ECCV, Fergus2006SIGGGRAPH, Shan2008ToG, Levin2009CVPR, Cho2009ToG} and those considering spatially varying blur~\cite{Whyte2010CVPR, Gupta2010ECCV, Tai2011PAMI}. Uniform deblurring methods assume that the blur kernel is identical for each pixel of the input image. Spatially varying deblurring methods assume that the blur kernels for each pixel may change with respect to its spatial location. Motion deblurring methods can also be classified into non-blind deblurring
\cite{Richardson1972,Krishnan2009NIPS, Tai2011PAMI} and blind deblurring \cite{Xu2010ECCV, Fergus2006SIGGGRAPH, Shan2008ToG, Levin2009CVPR, Cho2009ToG, Whyte2010CVPR, Gupta2010ECCV} methods. Non-blind deblurring methods assume a known blur kernel to recover the latent sharp image. In contrast, blind deblurring methods need to simultaneously recover both the latent image and the blur kernel. In this paper, we solve the challenging blind single image deblurring problem with spatially varying blur. 

\PAR{Optimization-based methods} rely on the blur formation model to recover the latent sharp image by minimizing an energy function \cite{Krishnan2009NIPS,Xu2010ECCV, Fergus2006SIGGGRAPH, Shan2008ToG, Levin2009CVPR, Cho2009ToG, Zheng2013ICCV}, \eg, using Gaussian \cite{Krishnan2009NIPS, Xu2010ECCV, Cho2009ToG} or Poisson \cite{Richardson1972, Tai2011PAMI} likelihood functions in the context of maximum-a-posteriori (MAP) estimation. Depending on the number of input blurry images, additional terms can be formulated by warping the other image(s) to a reference image using either dense flow or by combining relative camera poses with dense depth maps \cite{Kim2015CVPR,Park2017ICCV}. Due to the nonlinear and ill-posed (in the case of a single image) nature of the problem, prior information on either the motion blur kernel or the latent sharp image must be used to constrain the solution space \cite{Krishnan2009NIPS,Xu2010ECCV, Fergus2006SIGGGRAPH, Shan2008ToG,Krishnan2009NIPS, Xu2010ECCV, Cho2009ToG}.  

While optimization based approaches often offers better generalization performance, they are usually computational expensive, which prevents them from time constrained applications. We leverage the commonly used image formation model by optimization based approaches to construct a loss term for training our network in a self-supervised fashion. Since we optimize the parameters of our network once, the computationally complex optimization problem does not occur at test time, where we use the standard efficient feed-forward inference.

\PAR{Deep Learning based methods} use convolutional neural networks (CNNs) to recover the latent sharp image. Xu \etal \cite{Xu2014NIPS} propose a CNN with two sub-networks, a two-hidden-layer deconvolution network and a two-hidden-layer outlier rejection network. The network is trained end-to-end with known ground truth sharp images. An even deeper network with 15 layers was proposed in \cite{Hradis2015BMVC} for text image deblurring. \cite{Nah2017CVPR} further increased the number of layers to 40 in a multi-scale manner, resulting in a network with 120 layers for three scales. To further improve the network performance, an adversarial loss~\cite{Goodfellow2014NIPS} was used in \cite{Kupyn2018CVPR}. \cite{Tao2018CVPR,Zhang2018CVPR} use recurrent neural networks for single image deblurring. 

More recently, network architectures for multi-frame inputs, which are able to exploit temporal information, have been proposed~\cite{Su2017CVPR,Wieschollek2017ICCV,Kim2017ICCV}. Su \etal \cite{Su2017CVPR} uses a network with skip connections for video deblurring and Wieschollek \etal \cite{Wieschollek2017ICCV} exploit temporal information using a recurrent architecture. A spatio-temporal recurrent architecture with a small computational footprint was proposed in \cite{Kim2017ICCV}.

Deep learning-based approaches usually outperform optimization-based methods in terms of both efficiency and image quality. However, nearly all of the recent deep learning based methods are trained in a fully supervised manner with only few notable exceptions. Madam \etal \cite{Madam2018ECCV} adapted CycleGAN \cite{Zhu2017ICCV} to the single image deblurring task. Using unpaired sharp images for training, they obtained good performance in the specific domain of images (\eg, text, faces). In contrast to our approach, they require (unpaired) sharp images for training and perform poorly on blurry images ``in the wild'' as demonstrated by our experiments.
Chen \etal \cite{Chen2018ICCP} propose to include a self-consistency loss for supervision. However, they report that their self-supervised model leads to degenerated solutions and hence optimize a hybrid loss function which heavily relies on supervision in the form of sharp images. Furthermore, in contrast to our approach, their method requires triplets of blurry images for training and uses a memory and computational expensive look-up table for representing the blur kernel in a differentiable fashion. We avoid this problem using forward warping and differentiable mesh rendering.
\section{Method}
\label{sec:method}

\begin{figure*}[th!]
	\centering
	\includegraphics[width=\textwidth]{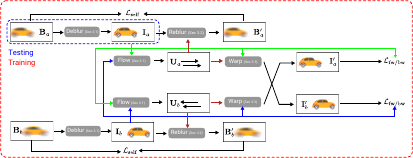}
	\caption{\textbf{Architecture of the proposed network.} Given two consecutive blurry images, $\bB_a$ and $\bB_b$, as input, our network computes the corresponding deblurred images, $\bI_a$ and $\bI_b$, as well as the bidirectional optical flow, $\bU_a$ and $\bU_b$. To self-supervise the training of the network, we construct a self-consistency photometric loss $\mathcal{L}_\text{self}$ and a forward-backward photometric consistency losses, $\mathcal{L}_\text{fw/bw}$.}
	\label{fig_network}
\end{figure*}

\figref{fig_network} shows the overall architecture of our model. It comprises four main parts, \ie, the DeblurNet, the FlowNet, the reblur block, and image warping. The DeblurNet is used for single image motion deblurring. It accepts a single blurry image as input and outputs the corresponding sharp image. The two deblurred images are then fed into the FlowNet to estimate a bi-directional dense optical flow field, which will be used to compute spatially varying blur kernels for each blurry image. Given the estimated blur kernels, we reblur the latent sharp image to form a self-consistency loss to supervise the training of our network. The FlowNet is trained by maximizing cross-view photometric consistency, which is estimated by image warping. While our approach uses two images for training, the DeblurNet only uses a single input image. After training, our method can thus be used for single image deblurring. The whole network is trained end-to-end without using any ground truth data in the form of sharp images or optical flow. We will now present all components of our model (\ie, the deblurring, optical flow, reblurring and image warping components as well as the loss functions) in detail.

\subsection{Deblurring and Optical Flow}\label{sec_deblurNet}

For the deblurring and optical flow modules, we take advantage of existing neural network architectures which have performed well in the past for the respective supervised learning tasks \cite{Nah2017CVPR, Kupyn2018CVPR, Tao2018CVPR, Sun2018CVPR, Ilg2017CVPR}. In particular, we adopt the single image deblurring network from Tao \etal \cite{Tao2018CVPR} and the dense optical flow estimation network PWC-Net from Sun \etal \cite{Sun2018CVPR}. We make the following modifications for the deblurring network for our particular problem: 1) We replace the deconvolution layer with bilinear upsampling followed by a 3x3 convolution to avoid upsampling artifacts. 2) We add one more Encoder-Decoder block to increase the capacity of the network. 3) We train the network at a single scale without using the LSTM layer to improve both the training and test efficiency. The resulting network is more efficient than the original network while keeping similar deblurring performance.

\subsection{Reblurring}

The reblurring module encapsulates the physical image formation process, which blurs a sharp image based on the optical flow.
Digital cameras operate by collecting photons during the time of exposure and converting those into measurable charge.
This process can be formalized by considering the blurred color image $\bB\in\nR^{W \times H \times 3}$ as the result of integrating virtual sharp images $\bI_t\in\nR^{W \times H \times 3}$:

\begin{equation} \label{eq_formation_model}
\bB(\bx) = \int_{0}^{\tau}\bI_t(\bx) \text{dt} \approx \frac{1}{2N+1} \sum_{i=-N}^{N} \bI_{i}(\bx) 
\end{equation}
Here, $\tau$ is the exposure time, $\bx\in\nR^2$ represents the pixel location, $\bB(\bx)$ denotes the motion blurred image at pixel $\bx$, and $\bI_t(\bx)$ is the virtual sharp image at pixel $\bx$ and time $t$. The continuous integration can be approximated by using a finite sample size of $2N+1$ virtual sharp frames $\bI_i$. We denote the central reference frame, which is the latent sharp image to be estimated, as $\bI_0$.

As the exposure time $\tau$ is typically small ($<$200 ms), we may assume that during the time of exposure the image content is primarily affected by image motion and not by other changes like object appearance or illumination. We thus model the virtual sharp frames $\bI_i$ as the result of the sharp central reference frame $\bI_0$ warped by optical flow $\bu_{i\rightarrow 0}$:
\begin{equation}
\bI_{i}(\bx) = \bI_{0}(\bx + \bu_{i \rightarrow 0})
\end{equation}
Here, $\bu_{i \rightarrow 0}\in\nR^2$ denotes the optical flow from virtual image $\bI_i$ to reference image $\bI_0$ at pixel $\bx$. Thus, we can reformulate \eqref{eq_formation_model} as 
\begin{equation}
 \bB(\bx) \approx \frac{1}{2N+1} \sum_{i=-N}^{N} \bI_0(\bx+\bu_{i\rightarrow 0}) 
 \label{eq:image_formation_flow}
\end{equation}
and estimate $\bI_0$ as well as the optical flow fields $\bU_{i \rightarrow 0}$ instead of all virtual frames $\bI_i$ for solving the deblurring problem. However, the problem is still severely underconstrained as we would need to estimate one optical flow per frame $i\in\{-N,\dots,N\}$.

We therefore further simplify the model by assuming linear motion during the time of exposure.
This is a reasonable assumption in many scenarios where the exposure time is comparably small and rapid motion changes during this time are prevented by the mass and inertia of physical objects (\eg, when the camera is mounted to a vehicle). 

Let $\bu\in\nR^2$ denote the optical flow from frame $\bI_0$ to frame $\bI_1$ at pixel $\bx$. 
Assuming linear motion and equidistant time steps, we obtain the optical flow from frame $0$ to frame $i$ as $\bu_{0\rightarrow i} = i \cdot \bu$. Note that in this model the direction of the optical flow is reversed compared to \eqref{eq:image_formation_flow}. We must therefore apply forward warping to obtain the virtual sharp images $\bI_i$. This yields
\begin{equation}
\bB(\bx) \approx \frac{1}{2N+1} \sum_{i=-N}^{N} (\cW_{0 \rightarrow i} \circ \bI_0)(\bx) \enspace
\label{eq:image_formation_flow_linear}
\end{equation}
where the operator $\cW_{0 \rightarrow i}$ warps the reference frame $\bI_0$ into the virtual frame $\bI_i$ based on the interpolated flow $\bu_{0\rightarrow i}$.
We next describe our implementation of the forward warping operator $\cW_{0\rightarrow i}$. Note that this operator needs to be differentiable as both the reference frame $\bI_0$ and the optical flow $\bU$ are outputs of neural networks.

\begin{figure}[t!]
	\centering
	\includegraphics[width=\columnwidth]{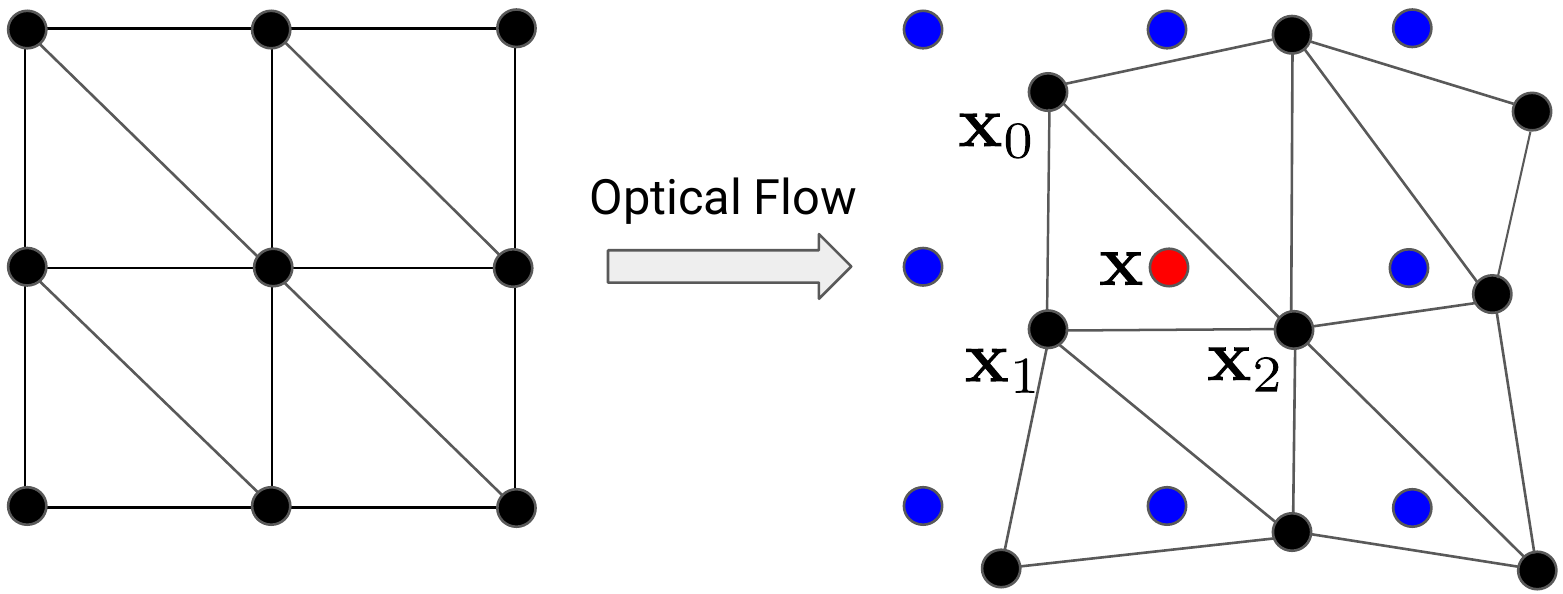}
	\caption{\textbf{Differentiable forward warping.} We construct a regular triangular lattice from the pixel grid of the reference image $\bI_0$ (left). We then warp each vertex of this lattice according to the optical flow $\bu_{0\rightarrow i}$. The intensities of $\bI_i$ (\ie, of the blue pixels) are obtained by linear interpolation.}
	\label{fig_forward_warping}
\end{figure}

We first construct a regular triangular lattice from the pixel grid by connecting vertices from adjacent pixels as shown in \figref{fig_forward_warping} for an example image of size $3 \times 3$ pixels. We then warp each vertex of this lattice according to the optical flow $\bu_{0 \rightarrow i}$. The intensities of $\bI_i$ (\ie, of the blue pixels) are obtained by linear interpolation \footnote{Note that bilinear interpolation cannot be applied since the warped grid might not be rectangular due to the non-uniform optical flow, as shown in \figref{fig_forward_warping}.}. Consider the red pixel  $\bx$ in \figref{fig_forward_warping} 
as an example. Let further $\bx_0$, $\bx_1$ and $\bx_2$ denote the positions of the vertices belonging to the triangle which covers the red pixel. Then, $\bI_i$ is obtained as
\begin{equation}
 \bI_i(\bx) = \omega_0 \bI_0(\bx_0) + \omega_1 \bI_0(\bx_1) + \omega_2 \bI_0(\bx_2) \enspace
\end{equation} 
where $\omega_0$, $\omega_1$ and $\omega_2$ denote the barycentric coordinates of the point in the triangle. The synthesized motion blurred image can then be computed as the average of all the warped frames as described in \eqref{eq_formation_model}. In the case of occlusions, \ie, when multiple triangles overlap a single pixel, we consider the triangle with the largest motion to be in front. This is a commonly used  heuristics \cite{Kato2018CVPR} which often holds true in practice (in particular when image motion is dominated by camera motion). Note that due to the linear interpolation, the warping function $\cW_{0 \rightarrow i}$ is piecewise smooth. As illustrated in \figref{fig_network}, our model is symmetric, thus we reblur both the first and the second frame and compare the reblurred result to the original blurry images using a photoconsistency loss.
We will use $\bB'$ to denote the reblurred image and $\bB$ to denote the blurred input image in the following.

\subsection{Image Warping}

We found that a photoconsistency loss on the reblurred images alone is insufficient to constrain the optical flow. We thus add an additional self-supervised photometric loss on the optical flow as proposed in prior work \cite{Janai2018ECCV,Meister2018AAAI,Wang2018CVPR} and detailed in \secref{sec:method_loss_functions}. The input to this loss function is the deblurred image and the deblurred image from the other frame warped based on the estimated optical flow. To warp the images into each other, we exploit backward warping as the optical flow in both directions is known. Let $\bI_a\equiv\bI^a_0$ and $\bI_b\equiv\bI^b_0$ denote the first and the second deblurred image, and let $\bu_{a \rightarrow b}$ and $\bu_{b \rightarrow a}$ denote the optical flow between them\footnote{Note that the flow $\bu_{a\rightarrow b}$ / $\bu_{b \rightarrow a}$ and $\bu$ from the previous section are related by a known constant that depends on the frame rate, exposure time and $N$.}.
The warped deblurred images are obtained as 
\begin{align}
 \bI'_a(\bx) &= \bI_b(\bx + \bu_{a \rightarrow b}) \\
 \bI'_b(\bx) &= \bI_a(\bx + \bu_{b \rightarrow a}) 
\end{align}
using bilinear interpolation \cite{Jaderberg2015NIPS}. Note that no triangular mesh needs to be constructed during backward warping.

\subsection{Loss Functions}
\label{sec:method_loss_functions}

Our network comprises two types of losses: a self-consistency loss $\cL_\text{self}$ and a forward-backward consistency loss $\cL_\text{fw/bw}$.
The self-consistency loss 
\begin{equation}
\cL_\text{self} = {\Vert\bB'_a-\bB_a\Vert}_1 + {\Vert\bB'_b-\bB_b\Vert}_1
\label{eq:loss_self}
\end{equation}
penalizes differences between the synthesized motion blurred images $\bB'_a$, $\bB'_b$ and the original blurred inputs $\bB_a$, $\bB_b$ 
using a $\ell_1$ loss function.
Similarly, the forward-backward consistency loss 
\begin{equation}
\cL_\text{fw/bw} = {\Vert\bI'_a-\bI_a\Vert}_1 + {\Vert\bI'_b-\bI_b\Vert}_1 
\label{eq:loss_fwbw}
\end{equation}
penalizes differences between the warped deblurred images $\bI'_a$, $\bI'_b$ and the estimated deblurred images $\bI_a$, $\bI_b$. 
The final loss is a weighted combination 
\begin{align}\label{eq:loss_sum}
\cL&=\cL_\text{self} + \lambda \cL_\text{fw/bw},
\end{align}
where $\lambda$ is a hyper-parameter to balance both losses.

\subsection{Occlusion Handling}
As occlusions affect the training of our network, especially at image boundaries, we detect occluded image regions and mask the loss functions \eqref{eq:loss_self} and \eqref{eq:loss_fwbw} accordingly. We follow the method used in \cite{Wang2018CVPR} to detect occluded image regions. More specifically, we compute the non-occluded regions in $\bI_a$ by following the optical flow $\bU_{b\rightarrow a}$ from $\bI_b$ to $\bI_a$. We consider all pixels of $\bI_a$ which can be reached from $\bI_b$ via $\bU_{b\rightarrow a}$ as non-occluded. Similarly, we can also compute a mask for each virtual frame by following the optical flow from the central image to the virtual frame $\bu_{0\rightarrow i}$. Since the synthesized blurry image is computed as the average of these virtual frames, we compute the final mask for $\cL_{self}$ as the product of all masks for the virtual frames.

\subsection{Differences with the method proposed by \cite{Chen2018ICCP}} \label{sec_diff_to_chen}
The overall structure of our method is similar to the work from \cite{Chen2018ICCP}. However, we are different in the following two key aspects: \textbf{1)} In order to achieve state-of-the-art performance, \cite{Chen2018ICCP} uses a supervised loss (section 3.4 from \cite{Chen2018ICCP}). \cite{Chen2018ICCP} is thus actually a supervised method. \textbf{2)} The core component of both methods, i.e., the reblurring module, is different. In fact, blurring a sharp image using convolutions as done in \cite{Chen2018ICCP} is physically incorrect (section 3.3 from \cite{Chen2018ICCP}) and only holds for spatially uniform blur. We will make the differences to \cite{Chen2018ICCP} more clear.

For simplicity, let us assume we have a one dimensional sharp image $\mathbf{I}$ with $N$ pixels. We further assume the blur kernel corresponds to pixel $\mathbf{I}_i$ as $\{-2, 2\}$ in the form of bidirectional 1D flow. Using the definition from \cite{Chen2018ICCP}, the convolution based model results in a blurred image of $\mathbf{I}_i$ as $\mathbf{B}_i=\frac{1}{5}\sum_{j=-2}^{2} \mathbf{I}_{i+j}$. A blur kernel $\{-2, 2\}$ of $\mathbf{I}_i$ means that $\mathbf{I}_i$ will contribute to $\mathbf{B}_{i-2}, \mathbf{B}_{i-1},\mathbf{B}_i, \mathbf{B}_{i+1}, \mathbf{B}_{i+2}$ physically, in contrast to that $\mathbf{I}_{i-2}, \mathbf{I}_{i-1},\mathbf{I}_i, \mathbf{I}_{i+1}, \mathbf{I}_{i+2}$ will contribute to $\mathbf{B}_i$ as what the convolution based model in \cite{Chen2018ICCP} does. Our model eliminates this problem by forward warping the sharp image $\mathbf{I}$ by a fraction of the blur kernels at each sampled timestamp. The blurred image is computed by averaging all these forward warped sharp images to simulate the real motion blurring image formation process. Experimental results shown later demonstrate that the algorithm relying on convolution based model exhibits ringing artifacts on egde boundaries, which degrade the deblurred images.


\section{Experimental Evaluation}
\label{sec:results}

\boldparagraph{Datasets}
The dataset from \cite{Nah2017CVPR} is commonly used to benchmark single image motion deblurring algorithms. It is collected from a hand-held camera. Stronger blur was artificially created by shaking the camera during the recordings. It results in very non-linear camera motions, which violates our motion assumption. Therefore, we collected a new large dataset using a professional Fastec TS5\footnote{https://www.fastecimaging.com/fastec-high-speed-cameras-ts-series/} high speed camera mounted on a car. The dataset consists of 196 sequences in total, which are collected at 1200 fps with VGA resolution in diverse environments. The motion blurred images are generated by averaging several consecutive frames (\ie, 1$\sim$50 frames) to simulate the real physical image formation process. To reduce the redundancy per image sequence, we limit the maximum number of blurry-sharp image pairs to 20 per sequence, which results in a total of 3606 pairs. We split the dataset into 157 training sequences and 39 test sequences, which results in 2820 image pairs for training and 786 image pairs for evaluation.

We also collect a real motion blurry dataset with 2302 images. The camera is mounted on a tram and captures images at around 50 FPS with a resolution of 752$\times$480 pixels. The dataset is collected at late afternoon and night, when the motion blur would really occur. We split 2062 images for training and 240 images for test.

\boldparagraph{Implementation details}
We implemented our network in PyTorch~\cite{paszke2017automatic}. We empirically set the hyper-parameter $\lambda$ to be $2.0$. To better initialize the network, we pretrain both the DeblurNet and PWC-Net on the blurry images. In particular, we pretrain the DeblurNet for 30 epochs to learn the identity mapping from blurry image to blurry image. We pretrain the PWC-Net for 200 epochs with the blurry sequences in a self-supervised manner. The learning rate used for both networks is $10^{-4}$. The whole network is then trained jointly for another 500 epochs, with a learning rate of $10^{-4}$ for the first 260 epochs and then decayed by half every 40 epochs.

\boldparagraph{Baselines and experimental settings}
We compare the single image deblurring results of our network quantitatively and qualitatively with a state-of-the-art optimization-based method \cite{Xu2013CVPR}, supervised methods \cite{Nah2017CVPR, Tao2018CVPR, Kupyn2018CVPR} as well as the domain specific self-supervised method from \cite{Madam2018ECCV}. We train all networks with their recommended hyperparameter settings on our synthetic dataset. For the optimization-based method from \cite{Xu2013CVPR}, we increase the blur kernel size to 10 pixels to account for the large motions present in our dataset.

\boldparagraph{Evaluation metrics}
We use the Peak Signal to Noise Ratio (PSNR) and the Structural Similarity Index (SSIM) measures commonly used in the community \cite{Shan2008ToG, Nah2017CVPR, Tao2018CVPR} to evaluate the quality of the deblurring results. Larger PSNR/SSIM values indicate better image quality. The efficiency of the methods is evaluated by their total time consumption, but excluding the image loading and saving time. 

\boldparagraph{Ablation studies on the modified DeblurNet architecture}
As discussed in Sec.\ref{sec_deblurNet}, we did several improvements to the original network from \cite{Tao2018CVPR}. To evaluate the efficacy of the new network, we train both networks in a supervised manner on our synthetic dataset. \tabref{Tb_quantative_network} presents the comparisons when evaluated under the same settings. It demonstrates our DeblurNet is more efficient than the original network while has slightly better deblurring performance.

\begin{table}
	\centering
	\small{
	\begin{tabular}{lccr} 
	\toprule
	Network & PSNR$\uparrow$ & SSIM$\uparrow$  & Time $\downarrow$\\ 
	\midrule
	SRN-Deblur \cite{Tao2018CVPR}  & 34.64 dB  & 0.93 & 0.13 s\\
	\midrule
	Ours (supervised)  & 35.04 dB  & 0.94 & 0.05 s\\
	\bottomrule
	\end{tabular}
	}
	\caption{\textbf{Single image deblurring comparison on the synthetic dataset.} We compare our modified network with the original network from Tao et al. \cite{Tao2018CVPR} by training both in a supervised way.}
	\label{Tb_quantative_network}
\end{table}

\boldparagraph{Ablation studies on the self-supervision loss for the flow network}
To better understand the proposed algorithm, we perform an ablation study on the necessity to train the flow network in a self-supervised manner. We train our proposed network with and without the self-supervision loss for the flow network. The officially provided pretrained model on FlyingChair dataset \cite{ISKB18} is used if the self-supervision loss is disabled. Experimental results demonstrate that the flow network pretrained on the FlyingChair dataset \cite{ISKB18} can generalize to our dataset, but with limited performance. The resulting deblur network gives a PSNR metric as 31.23dB and a SSIM metric as 0.89 on our synthetic dataset, in contract to 32.24dB/0.91 if the network is trained in a fully self-supervised manner. It proves it is beneficial to train the flow network with a self-supervision loss.

\boldparagraph{Ablation studies on our proposed reblur model}
As discussed in Sec.\ref{sec_diff_to_chen}, our ablation study supports our claim about the difference between the reblurring modules. For fair comparisons, we trained both the network with convolution based reblur model and the network with our physically correct reblur model under the same settings in an unsupervised fashion. The network with convolution based image formation model yields a PSNR metric of 27.22dB and a SSIM metric of 0.8 on our synthetic dataset, while ours yields PSNR and SSIM metrics as 32.24dB and 0.91 respectively. 

\begin{figure*}
\centering
\minipage{0.33\textwidth}
  \includegraphics[width=\linewidth]{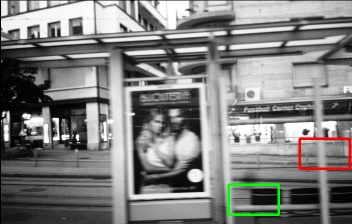}
\endminipage
\minipage{0.33\textwidth}
 \includegraphics[width=\linewidth]{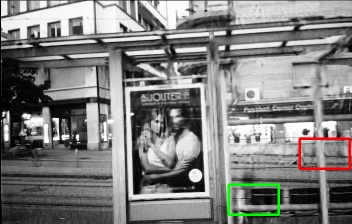}
\endminipage
\minipage{0.33\textwidth}
  \includegraphics[width=\linewidth]{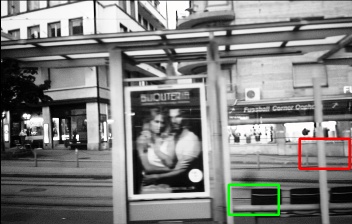}
\endminipage\vfill
\minipage{0.33\textwidth}
\includegraphics[width=\linewidth]{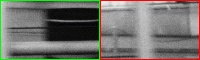}
\endminipage
\minipage{0.33\textwidth}
\includegraphics[width=\linewidth]{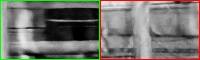}
\endminipage
\minipage{0.33\textwidth}
\includegraphics[width=\linewidth]{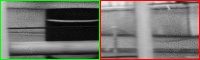}
\endminipage\vfill
\minipage{0.33\textwidth}
\includegraphics[width=\linewidth]{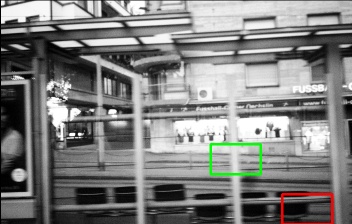}
\endminipage
\minipage{0.33\textwidth}
\includegraphics[width=\linewidth]{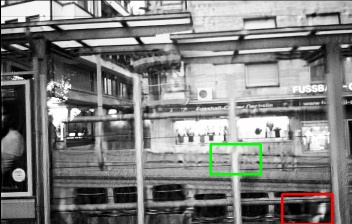}
\endminipage
\minipage{0.33\textwidth}
\includegraphics[width=\linewidth]{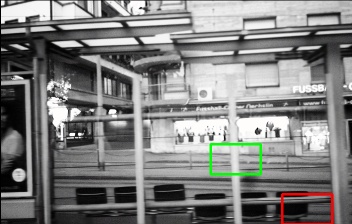}
\endminipage \vfill
\minipage{0.33\textwidth}
\includegraphics[width=\linewidth]{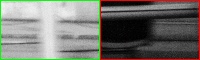}
\endminipage
\minipage{0.33\textwidth}
\includegraphics[width=\linewidth]{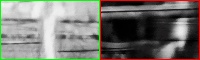}
\endminipage
\minipage{0.33\textwidth}
\includegraphics[width=\linewidth]{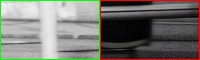}
\endminipage
\caption{\textbf{Qualitative evaluations on real dataset.} \textbf{First} Blurry image. \textbf{Second} Deblurred image by the official pretrained network from Tao \etal \cite{Tao2018CVPR}. \textbf{Third} Deblurred image by our method. The images are post processed for better visualization. Best viewed in digital version.}
\label{fig_qualative_results_real}
\vspace{-0.6cm}
\end{figure*}

\boldparagraph{The necessity to do self-supervised motion deblurring}
In real scenarios, motion blur usually occurs in bad illumination conditions. In these cases, it impedes the acquisition with low shutter times to obtain sharp images for supervised learning. One way to address this problem is to train a network with datasets collected under good illumination conditions and transfer the model to scenarios, where motion blur would occur. However, the generalization ability is still questionable due to the large difference between the image textures for both scenarios. 
We thus evaluate the generalization performance quantitatively and qualitatively with both our synthetic dataset and real dataset respectively. Note that it is not easy to obtain ground truth sharp images in real scenarios. We apply the pretrained networks on our test data directly, to evaluate the generalization performance. \tabref{Tb_quantative_deblur} and \figref{fig_qualative_results_real} present the experimental results. It demonstrates that all the baseline networks have limited generalization ability and perform worse than our method with a large margin. It proves that self-supervision is beneficial for the network to adapt to scenarios, where the ground truth data is difficult to obtain.

\begin{figure*}
\centering
\minipage{0.25\textwidth}
  \includegraphics[width=\linewidth]{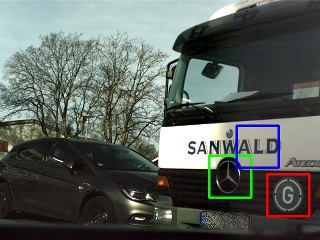}
\endminipage\hfill
\minipage{0.25\textwidth}
  \includegraphics[width=\linewidth]{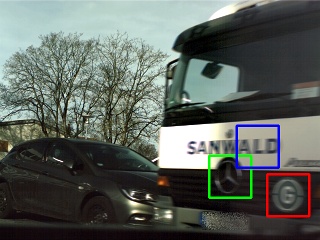}
\endminipage\hfill
\minipage{0.25\textwidth}
  \includegraphics[width=\linewidth]{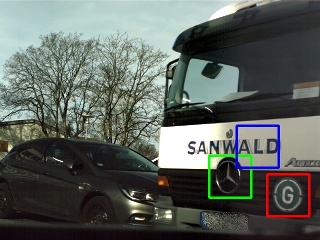}
\endminipage\hfill
\minipage{0.25\textwidth}
  \includegraphics[width=\linewidth]{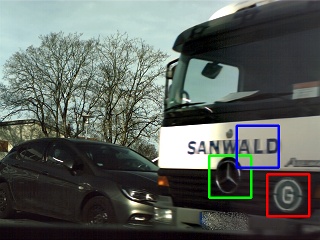}
\endminipage\vfill
\minipage{0.25\textwidth}
  \includegraphics[width=\linewidth]{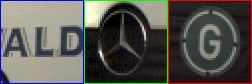}
\endminipage\hfill
\minipage{0.25\textwidth}
  \includegraphics[width=\linewidth]{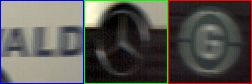}
\endminipage\hfill
\minipage{0.25\textwidth}
  \includegraphics[width=\linewidth]{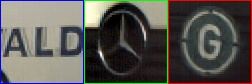}
\endminipage\hfill
\minipage{0.25\textwidth}
  \includegraphics[width=\linewidth]{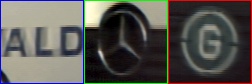}
\endminipage\vfill
\minipage{0.25\textwidth}
  \includegraphics[width=\linewidth]{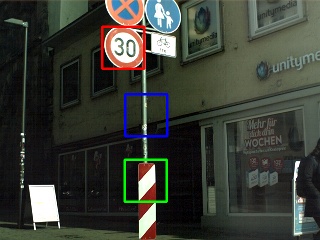}
\endminipage\hfill
\minipage{0.25\textwidth}
  \includegraphics[width=\linewidth]{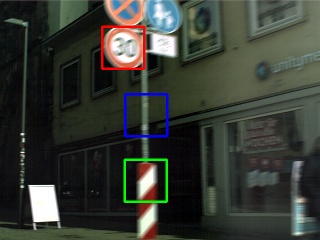}
\endminipage\hfill
\minipage{0.25\textwidth}
  \includegraphics[width=\linewidth]{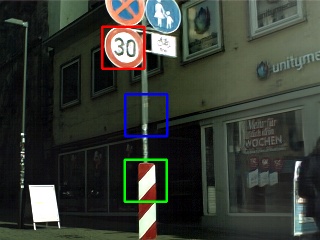}
\endminipage\hfill
\minipage{0.25\textwidth}
  \includegraphics[width=\linewidth]{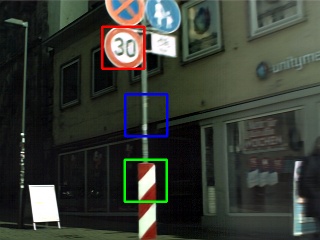}
\endminipage\vfill
\minipage{0.25\textwidth}
  \includegraphics[width=\linewidth]{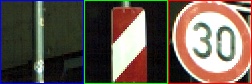}
\endminipage\hfill
\minipage{0.25\textwidth}
  \includegraphics[width=\linewidth]{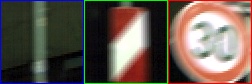}
\endminipage\hfill
\minipage{0.25\textwidth}
  \includegraphics[width=\linewidth]{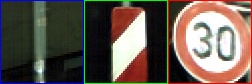}
\endminipage\hfill
\minipage{0.25\textwidth}
  \includegraphics[width=\linewidth]{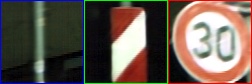}
\endminipage\vfill
\caption{\textbf{Qualitative comparisons on synthetic dataset.} \textbf{First}: Ground truth sharp image. \textbf{Second}: Input blurry image. \textbf{Third}: Deblurring results of the supervised method from Tao et al. \cite{Tao2018CVPR}; The network is retrained on our dataset. \textbf{Fourth}: Deblurring results of the proposed self-supervised learning method. }
\label{fig_qualative_results}
\end{figure*}

\boldparagraph{Quantitative and qualitative evaluations on synthetic dataset}
\tabref{Tb_quantative_deblur} and \figref{fig_qualative_results} show quantitative and qualitative comparisons on the synthetic dataset. For the qualitative results, we only compare against the best supervised method \cite{Tao2018CVPR}. 
As can be seen in \tabref{Tb_quantative_deblur}, our method outperforms both \cite{Xu2013CVPR} and \cite{Madam2018ECCV} significantly in terms of PSNR and SSIM. For optimization-based single image deblurring algorithms (\eg, \cite{Xu2013CVPR}), they usually assume the motion blur is caused by either camera rotation or in-plane translation. However, this assumption is violated in our setting for a self-driving scenario. Thus, \cite{Xu2013CVPR} leads to poor performance on our dataset.
\cite{Madam2018ECCV} is designed for simple domain-specific blurry images, such as text and facial images. Therefore, it struggles on our dataset that exhibits complex real-world challenges which are harder to learn.
In comparison to supervised methods, our method demonstrates competitive results in our quantitative and qualitative evaluation. As expected, there is still a gap between our method and the supervised methods if the ground truth sharp images are available. However, our method outperforms them with a large margin if they are pretrained on other datasets. It demonstrates that self-supervision enables the network to generalize better to real scenarios, where the ground truth data is usually difficult to obtain. It also demonstrates that our method is amongst the fastest methods and can run in real time on a single GTX1080Ti Graphic card.

\begin{table}
	\centering
	\small{
	\begin{tabular}{lcccr} 
	\toprule
	& Method & PSNR$\uparrow$ & SSIM$\uparrow$  & Time $\downarrow$\\ 
	\midrule
	Opt.-based & Xu \etal \cite{Xu2013CVPR} & 26.04 dB & 0.78 & 377.8 s	\\ 
	\midrule
	           & DeepDeblur\cite{Nah2017CVPR} & 33.55 dB & 0.92 & 3.45 s \\
	Supervised & DeblurGAN\cite{Kupyn2018CVPR}& 33.23 dB & 0.91 & 0.06 s \\
	-retrained & SRN-Deblur\cite{Tao2018CVPR} & 34.64 dB & 0.93 & 0.13 s \\
    \midrule
               & DeepDeblur\cite{Nah2017CVPR} & 29.91 dB & 0.87 & 3.45 s \\
    Supervised & DeblurGAN\cite{Kupyn2018CVPR}& 28.70 dB & 0.88 & 0.06 s \\
    -pretrained& SRN-Deblur\cite{Tao2018CVPR} & 30.71 dB & 0.88 & 0.13 s \\
    \midrule
    self-      & Madam et al.\cite{Madam2018ECCV} & 21.69 dB & 0.75 & 0.25 s \\
    supervised & Ours                             & 32.24 dB & 0.91 & 0.05 s \\
    \bottomrule
	\end{tabular}
	}
	\caption{\textbf{Single image deblurring on synthetic dataset.} Supervised-retrained denotes we retrained the networks with our training data. Supervised-pretrained denotes we use the official pretrained models to evaluate on our test data directly.}
	\label{Tb_quantative_deblur}
\end{table}

\boldparagraph{Qualitative evaluations on real dataset}
Since we do not have ground truth sharp images in our real dataset, we cannot refine the baseline networks on it. We thus use the official pretrained networks for the experiments. \figref{fig_qualative_results_real} demonstrates that our method can successfully deblur the blurry images, while the pretrained network from Tao \etal \cite{Tao2018CVPR} results in images with artifacts. More experimental results can be found from our supplementary material at \href{https://github.com/ethliup/SelfDeblur}{https://github.com/ethliup/SelfDeblur}.

\section{Conclusion and Future Work}
\label{sec:conclusion}

In this paper, we have presented a self-supervised learning algorithm for image deblurring. Instead of using ground truth sharp images, we leverage the geometric constraints between two consecutive blurry images to supervise training of our network. Both the latent sharp image and motion blur kernel are estimated by a deblur network and an optical flow estimation nework, respectively. Experimental results show that the proposed algorithm outperforms the previously self-supervised method and can produce competitive results compared to supervised methods. It further demonstrates that our method can be trained with real motion blurry data and generalizes well to real unseen data. 

{\small
	\bibliographystyle{ieee}
	\bibliography{bibliography_long,bibliography}
}

\clearpage
\onecolumn
\begin{center}
	\textbf{\large Supplemental Materials: Self-Supervised Linear Motion Deblurring}
\end{center}

\setcounter{equation}{0}
\setcounter{figure}{0}
\setcounter{table}{0}
\setcounter{page}{1}
\setcounter{section}{0}
\makeatletter


\section{Introduction}
\label{sec:intro}
In this supplementary material, we present details on the relationship between $\bu_{a\rightarrow b}$/$\bu_{b\rightarrow a}$ (\ie, the optical flow between the latent sharp images) and $\bu$ (\ie, the optical flow between the central virtual frame and the first virtual frame), described in Section III.B and III.C of the main paper, the architecture of the deblurring network, as well as additional qualitative experimental results on single image deblurring.


\section{Relationship between $\bu_{a\rightarrow b}$/$\bu_{b \rightarrow a}$ and $\bu$}
\label{sec:flow}
In this section, we present the relationship between $\bu_{a\rightarrow b}$/$\bu_{b \rightarrow a}$ and $\bu$. $\bu_{a\rightarrow b}$ and $\bu_{b \rightarrow a}$ are the bidirectional dense optical flows between the latent sharp images $\bI_{a}$ and $\bI_{b}$, respectively. 
We assume the motion between $\bI_{a}$ and $\bI_{b}$ to be linear. Without loss of generality, we assume $\bu$ is used to synthesize the first blurry image $\bB_a$. Thus, we obtain $\bu$ as the flow from the central virtual frame $\bI_{0}$ to the first virtual image $\bI_1$ by linearly scaling $\bu_{a\rightarrow b}$ according to 
\begin{equation}
\bu \approx \frac{\tau_a}{2\text{N}\Delta t} \bu_{a\rightarrow b} \enspace ,
\end{equation}
where $\tau_a$ is the exposure time of $\bB_a$, $2\text{N}+1$ is the number of sampled virtual sharp frames to synthesize $\bB_a$, and $\Delta t$ is the time interval between $\bI_a$ and $\bI_b$. Similarly, if $\bu$ is used to synthesize the second blurry image $\bB_b$, we get 
\begin{equation}
\bu \approx \frac{\tau_b}{2\text{N}\Delta t} \bu_{b\rightarrow a} \enspace ,
\end{equation}
where $\tau_b$ is the exposure time of $\bB_b$.


\section{Architecture of the Deblurring Network}
\label{sec:network}

\begin{figure*}[th!]
	\centering
	\includegraphics[width=\textwidth]{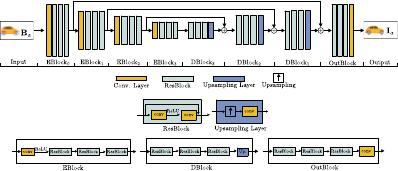}
	\caption{\textbf{Architecture of the deblurring network.} Given the input blurry image $\bB_a$, the deblurring network outputs the deblurred latent sharp image $\bI_a$. Best viewed in enlarged digital version.}
	\label{fig_network}
\end{figure*}

We adapt the single image deblurring network from Tao et al. \cite{Supp_Tao2018CVPR} for our approach. We make the following modifications for our particular problem: 1) We replace the deconvolution layer with bilinear upsampling followed by a 3x3 convolution to avoid upsampling artifacts. 2) We train the network at a single scale without using the LSTM layer to improve both the training and test efficiency. 3) We add one more Encoder-Decoder block to increase the capacity of the network. The details are shown in \figref{fig_network}.


\section{Additional Qualitative Experimental Results}
\label{sec:exp_fastec}
\begin{figure*}
	\centering
	\minipage{0.4\textwidth}
	\includegraphics[width=\linewidth]{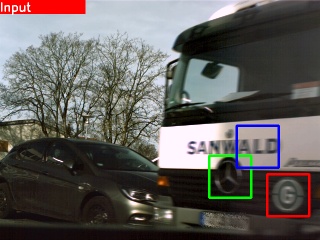}
	\endminipage\hfill
	\minipage{0.1\textwidth}
	\includegraphics[width=\linewidth]{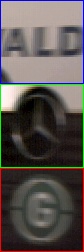}
	\endminipage\hfill
	\minipage{0.4\textwidth}
	\includegraphics[width=\linewidth]{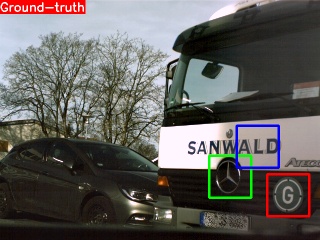}
	\endminipage\hfill
	\minipage{0.1\textwidth}
	\includegraphics[width=\linewidth]{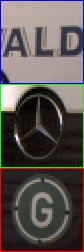}
	\endminipage\hfill
	\vspace{0.004\textheight}
	\minipage{0.4\textwidth}
	\includegraphics[width=\linewidth]{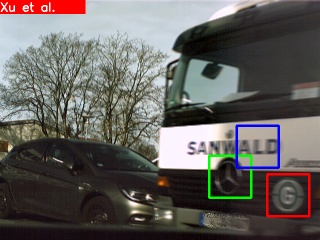}
	\endminipage\hfill
	\minipage{0.1\textwidth}
	\includegraphics[width=\linewidth]{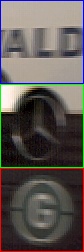}
	\endminipage\hfill
	\minipage{0.4\textwidth}
	\includegraphics[width=\linewidth]{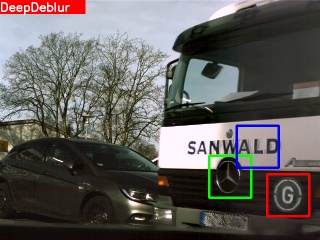}
	\endminipage\hfill
	\minipage{0.1\textwidth}
	\includegraphics[width=\linewidth]{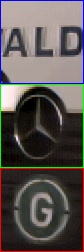}
	\endminipage\hfill
	\vspace{0.004\textheight}
	\minipage{0.4\textwidth}
	\includegraphics[width=\linewidth]{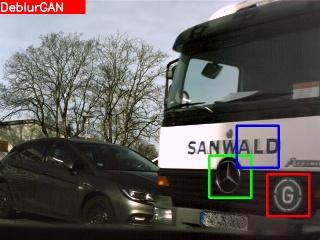}
	\endminipage\hfill
	\minipage{0.1\textwidth}
	\includegraphics[width=\linewidth]{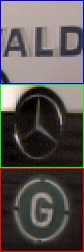}
	\endminipage\hfill
	\minipage{0.4\textwidth}
	\includegraphics[width=\linewidth]{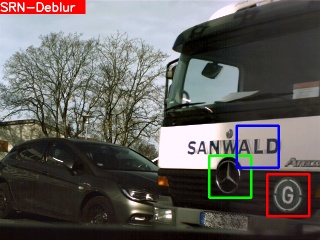}
	\endminipage\hfill
	\minipage{0.1\textwidth}
	\includegraphics[width=\linewidth]{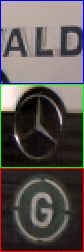}
	\endminipage\hfill
	\vspace{0.004\textheight}
	\minipage{0.4\textwidth}
	\includegraphics[width=\linewidth]{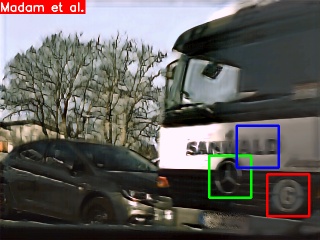}
	\endminipage\hfill
	\minipage{0.1\textwidth}
	\includegraphics[width=\linewidth]{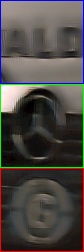}
	\endminipage\hfill
	\minipage{0.4\textwidth}
	\includegraphics[width=\linewidth]{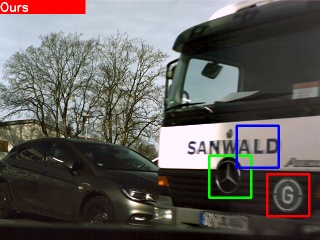}
	\endminipage\hfill
	\minipage{0.1\textwidth}
	\includegraphics[width=\linewidth]{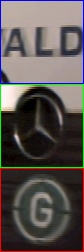}
	\endminipage\hfill
	\caption{\textbf{Qualitative comparisons on the Fastec dataset.} All the baseline networks are retrained on our Fastec dataset.}
	\label{fig_qualative_results_fastec_1}
	\vspace{0.6cm}
\end{figure*}

\begin{figure*}
	\centering
	\minipage{0.4\textwidth}
	\includegraphics[width=\linewidth]{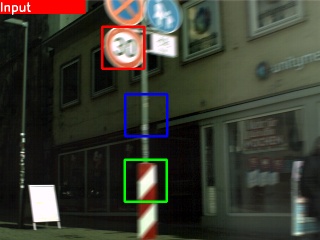}
	\endminipage\hfill
	\minipage{0.1\textwidth}
	\includegraphics[width=\linewidth]{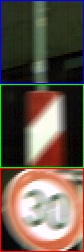}
	\endminipage\hfill
	\minipage{0.4\textwidth}
	\includegraphics[width=\linewidth]{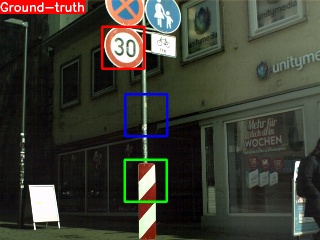}
	\endminipage\hfill
	\minipage{0.1\textwidth}
	\includegraphics[width=\linewidth]{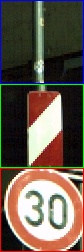}
	\endminipage\hfill
	\vspace{0.004\textheight}
	\minipage{0.4\textwidth}
	\includegraphics[width=\linewidth]{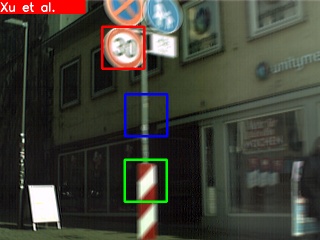}
	\endminipage\hfill
	\minipage{0.1\textwidth}
	\includegraphics[width=\linewidth]{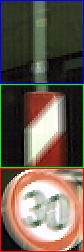}
	\endminipage\hfill
	\minipage{0.4\textwidth}
	\includegraphics[width=\linewidth]{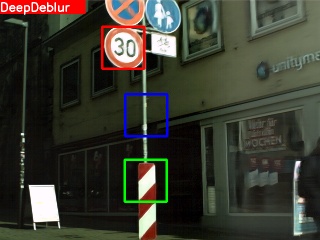}
	\endminipage\hfill
	\minipage{0.1\textwidth}
	\includegraphics[width=\linewidth]{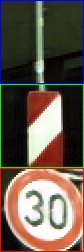}
	\endminipage\hfill
	\vspace{0.004\textheight}
	\minipage{0.4\textwidth}
	\includegraphics[width=\linewidth]{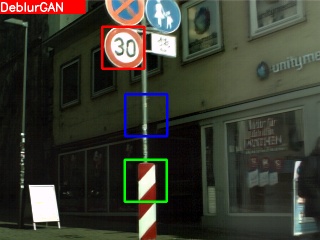}
	\endminipage\hfill
	\minipage{0.1\textwidth}
	\includegraphics[width=\linewidth]{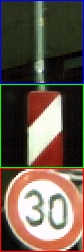}
	\endminipage\hfill
	\minipage{0.4\textwidth}
	\includegraphics[width=\linewidth]{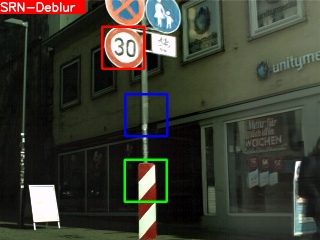}
	\endminipage\hfill
	\minipage{0.1\textwidth}
	\includegraphics[width=\linewidth]{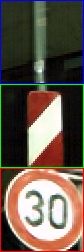}
	\endminipage\hfill
	\vspace{0.004\textheight}
	\minipage{0.4\textwidth}
	\includegraphics[width=\linewidth]{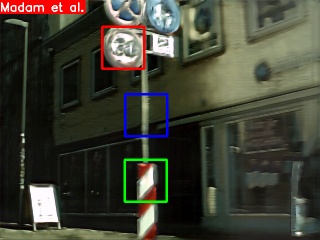}
	\endminipage\hfill
	\minipage{0.1\textwidth}
	\includegraphics[width=\linewidth]{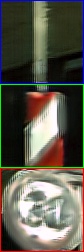}
	\endminipage\hfill
	\minipage{0.4\textwidth}
	\includegraphics[width=\linewidth]{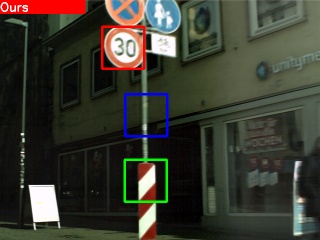}
	\endminipage\hfill
	\minipage{0.1\textwidth}
	\includegraphics[width=\linewidth]{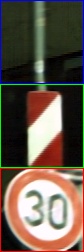}
	\endminipage\hfill
	\caption{\textbf{Qualitative comparisons on the Fastec dataset.} All the baseline networks are retrained on our Fastec dataset.}
	\label{fig_qualative_results_fastec_2}
	\vspace{0.6cm}
\end{figure*}

\begin{figure*}
	\centering
	\minipage{0.4\textwidth}
	\includegraphics[width=\linewidth]{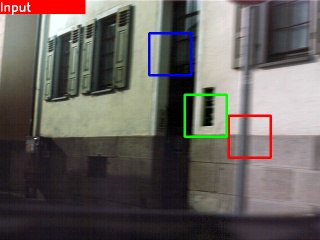}
	\endminipage\hfill
	\minipage{0.1\textwidth}
	\includegraphics[width=\linewidth]{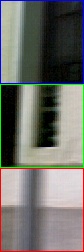}
	\endminipage\hfill
	\minipage{0.4\textwidth}
	\includegraphics[width=\linewidth]{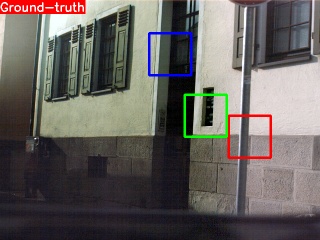}
	\endminipage\hfill
	\minipage{0.1\textwidth}
	\includegraphics[width=\linewidth]{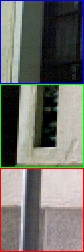}
	\endminipage\hfill
	\vspace{0.004\textheight}
	\minipage{0.4\textwidth}
	\includegraphics[width=\linewidth]{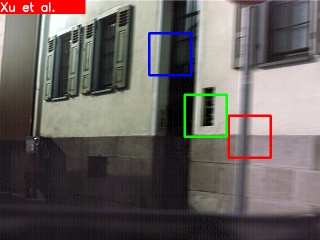}
	\endminipage\hfill
	\minipage{0.1\textwidth}
	\includegraphics[width=\linewidth]{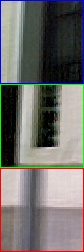}
	\endminipage\hfill
	\minipage{0.4\textwidth}
	\includegraphics[width=\linewidth]{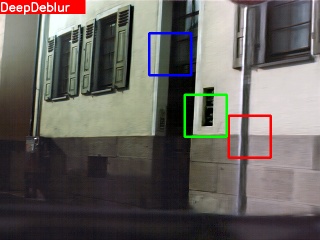}
	\endminipage\hfill
	\minipage{0.1\textwidth}
	\includegraphics[width=\linewidth]{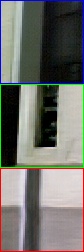}
	\endminipage\hfill
	\vspace{0.004\textheight}
	\minipage{0.4\textwidth}
	\includegraphics[width=\linewidth]{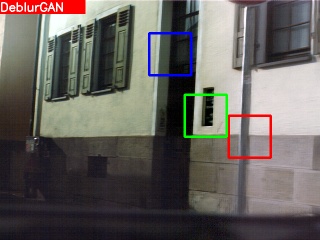}
	\endminipage\hfill
	\minipage{0.1\textwidth}
	\includegraphics[width=\linewidth]{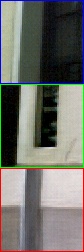}
	\endminipage\hfill
	\minipage{0.4\textwidth}
	\includegraphics[width=\linewidth]{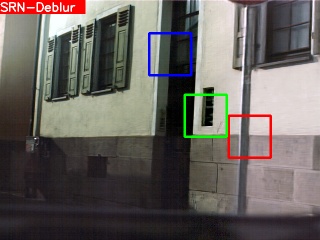}
	\endminipage\hfill
	\minipage{0.1\textwidth}
	\includegraphics[width=\linewidth]{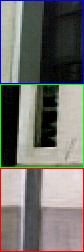}
	\endminipage\hfill
	\vspace{0.004\textheight}
	\minipage{0.4\textwidth}
	\includegraphics[width=\linewidth]{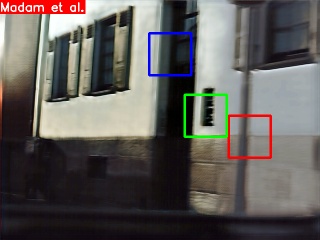}
	\endminipage\hfill
	\minipage{0.1\textwidth}
	\includegraphics[width=\linewidth]{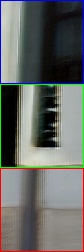}
	\endminipage\hfill
	\minipage{0.4\textwidth}
	\includegraphics[width=\linewidth]{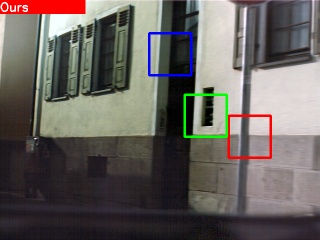}
	\endminipage\hfill
	\minipage{0.1\textwidth}
	\includegraphics[width=\linewidth]{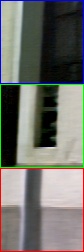}
	\endminipage\hfill
	\caption{\textbf{Qualitative comparisons on the Fastec dataset.} All the baseline networks are retrained on our Fastec dataset.}
	\label{fig_qualative_results_fastec_3}
	\vspace{0.6cm}
\end{figure*}

\begin{figure*}
	\centering
	\minipage{0.3976\textwidth}
	\includegraphics[width=\linewidth]{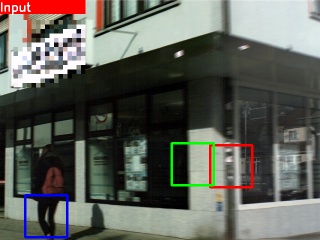}
	\endminipage\hfill
	\minipage{0.0994\textwidth}
	\includegraphics[width=\linewidth]{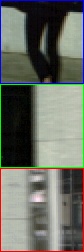}
	\endminipage\hfill
	\minipage{0.3976\textwidth}
	\includegraphics[width=\linewidth]{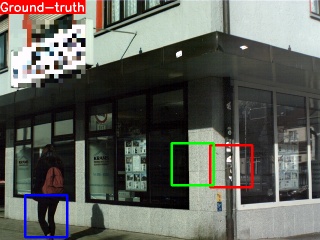}
	\endminipage\hfill
	\minipage{0.0994\textwidth}
	\includegraphics[width=\linewidth]{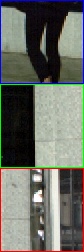}
	\endminipage\hfill
	\vspace{0.004\textheight}
	\minipage{0.3976\textwidth}
	\includegraphics[width=\linewidth]{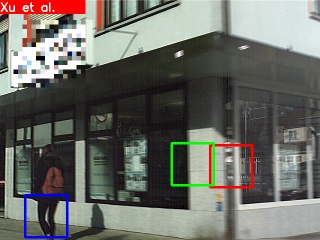}
	\endminipage\hfill
	\minipage{0.0994\textwidth}
	\includegraphics[width=\linewidth]{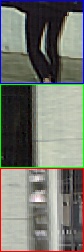}
	\endminipage\hfill
	\minipage{0.3976\textwidth}
	\includegraphics[width=\linewidth]{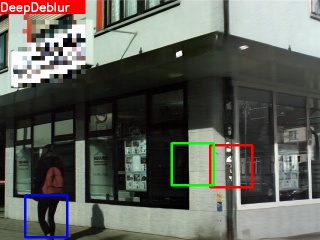}
	\endminipage\hfill
	\minipage{0.0994\textwidth}
	\includegraphics[width=\linewidth]{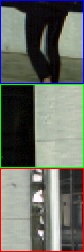}
	\endminipage\hfill
	\vspace{0.004\textheight}
	\minipage{0.3976\textwidth}
	\includegraphics[width=\linewidth]{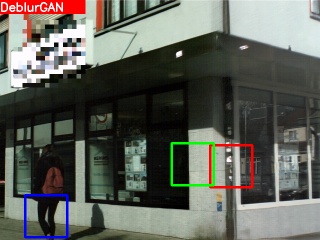}
	\endminipage\hfill
	\minipage{0.0994\textwidth}
	\includegraphics[width=\linewidth]{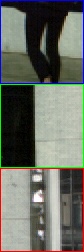}
	\endminipage\hfill
	\minipage{0.3976\textwidth}
	\includegraphics[width=\linewidth]{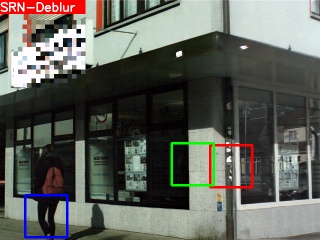}
	\endminipage\hfill
	\minipage{0.0994\textwidth}
	\includegraphics[width=\linewidth]{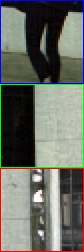}
	\endminipage\hfill
	\vspace{0.004\textheight}
	\minipage{0.3976\textwidth}
	\includegraphics[width=\linewidth]{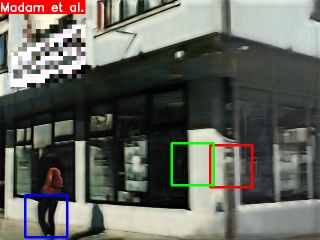}
	\endminipage\hfill
	\minipage{0.0994\textwidth}
	\includegraphics[width=\linewidth]{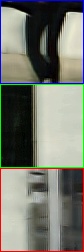}
	\endminipage\hfill
	\minipage{0.3976\textwidth}
	\includegraphics[width=\linewidth]{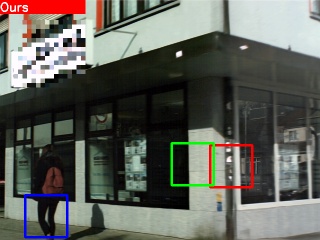}
	\endminipage\hfill
	\minipage{0.0994\textwidth}
	\includegraphics[width=\linewidth]{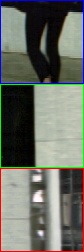}
	\endminipage\hfill
	\caption{\textbf{Qualitative comparisons on the Fastec dataset.} All the baseline networks are retrained on our Fastec dataset.}
	\label{fig_qualative_results_fastec_4}
	\vspace{0.6cm}
\end{figure*}

\begin{figure*}
	\centering
	\minipage{0.3976\textwidth}
	\includegraphics[width=\linewidth]{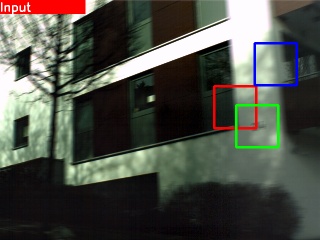}
	\endminipage\hfill
	\minipage{0.0994\textwidth}
	\includegraphics[width=\linewidth]{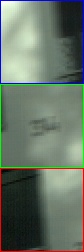}
	\endminipage\hfill
	\minipage{0.3976\textwidth}
	\includegraphics[width=\linewidth]{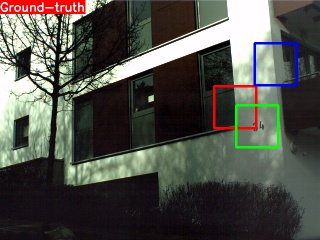}
	\endminipage\hfill
	\minipage{0.0994\textwidth}
	\includegraphics[width=\linewidth]{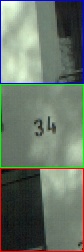}
	\endminipage\hfill
	\vspace{0.004\textheight}
	\minipage{0.3976\textwidth}
	\includegraphics[width=\linewidth]{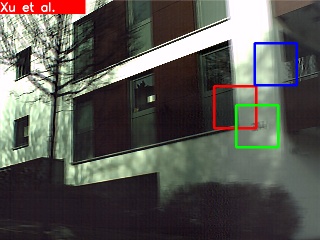}
	\endminipage\hfill
	\minipage{0.0994\textwidth}
	\includegraphics[width=\linewidth]{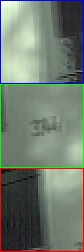}
	\endminipage\hfill
	\minipage{0.3976\textwidth}
	\includegraphics[width=\linewidth]{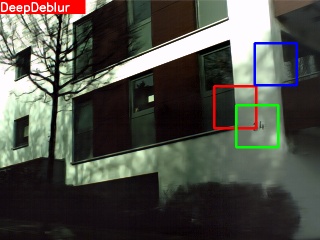}
	\endminipage\hfill
	\minipage{0.0994\textwidth}
	\includegraphics[width=\linewidth]{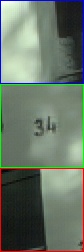}
	\endminipage\hfill
	\vspace{0.004\textheight}
	\minipage{0.3976\textwidth}
	\includegraphics[width=\linewidth]{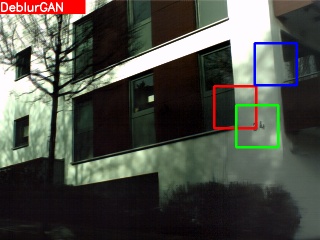}
	\endminipage\hfill
	\minipage{0.0994\textwidth}
	\includegraphics[width=\linewidth]{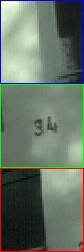}
	\endminipage\hfill
	\minipage{0.3976\textwidth}
	\includegraphics[width=\linewidth]{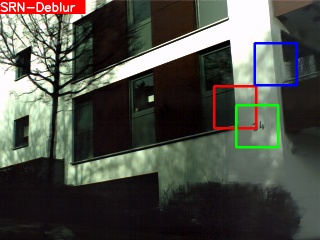}
	\endminipage\hfill
	\minipage{0.0994\textwidth}
	\includegraphics[width=\linewidth]{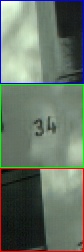}
	\endminipage\hfill
	\vspace{0.004\textheight}
	\minipage{0.3976\textwidth}
	\includegraphics[width=\linewidth]{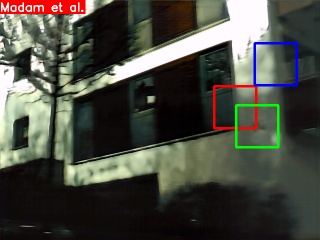}
	\endminipage\hfill
	\minipage{0.0994\textwidth}
	\includegraphics[width=\linewidth]{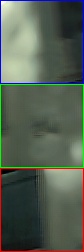}
	\endminipage\hfill
	\minipage{0.3976\textwidth}
	\includegraphics[width=\linewidth]{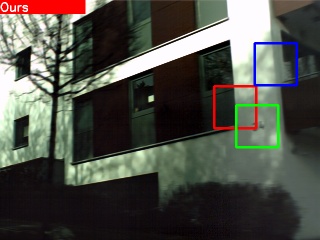}
	\endminipage\hfill
	\minipage{0.0994\textwidth}
	\includegraphics[width=\linewidth]{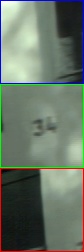}
	\endminipage\hfill
	\caption{\textbf{Qualitative comparisons on the Fastec dataset.} All the baseline networks are retrained on our Fastec dataset.}
	\label{fig_qualative_results_fastec_5}
	\vspace{0.6cm}
\end{figure*}

\begin{figure*}
	\centering
	\minipage{0.3976\textwidth}
	\includegraphics[width=\linewidth]{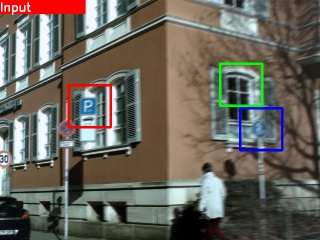}
	\endminipage\hfill
	\minipage{0.0994\textwidth}
	\includegraphics[width=\linewidth]{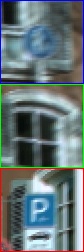}
	\endminipage\hfill
	\minipage{0.3976\textwidth}
	\includegraphics[width=\linewidth]{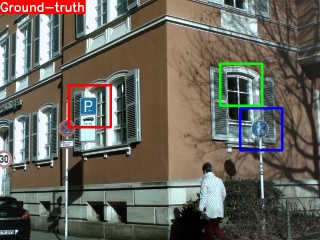}
	\endminipage\hfill
	\minipage{0.0994\textwidth}
	\includegraphics[width=\linewidth]{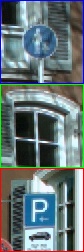}
	\endminipage\hfill
	\vspace{0.004\textheight}
	\minipage{0.3976\textwidth}
	\includegraphics[width=\linewidth]{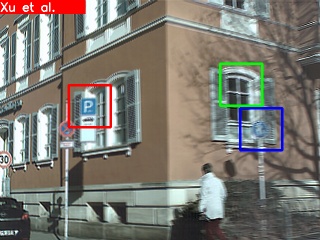}
	\endminipage\hfill
	\minipage{0.0994\textwidth}
	\includegraphics[width=\linewidth]{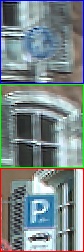}
	\endminipage\hfill
	\minipage{0.3976\textwidth}
	\includegraphics[width=\linewidth]{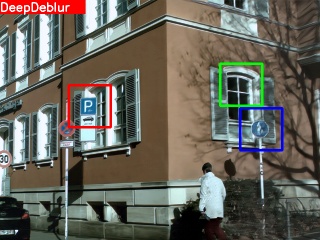}
	\endminipage\hfill
	\minipage{0.0994\textwidth}
	\includegraphics[width=\linewidth]{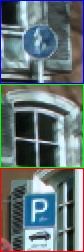}
	\endminipage\hfill
	\vspace{0.004\textheight}
	\minipage{0.3976\textwidth}
	\includegraphics[width=\linewidth]{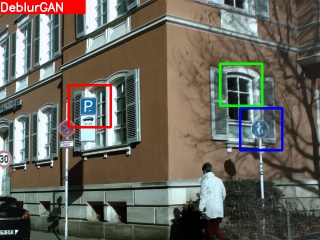}
	\endminipage\hfill
	\minipage{0.0994\textwidth}
	\includegraphics[width=\linewidth]{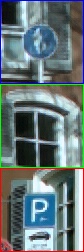}
	\endminipage\hfill
	\minipage{0.3976\textwidth}
	\includegraphics[width=\linewidth]{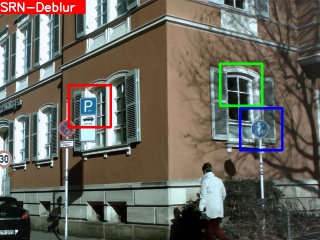}
	\endminipage\hfill
	\minipage{0.0994\textwidth}
	\includegraphics[width=\linewidth]{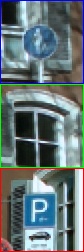}
	\endminipage\hfill
	\vspace{0.004\textheight}
	\minipage{0.3976\textwidth}
	\includegraphics[width=\linewidth]{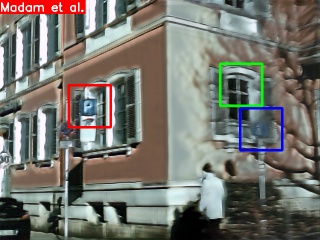}
	\endminipage\hfill
	\minipage{0.0994\textwidth}
	\includegraphics[width=\linewidth]{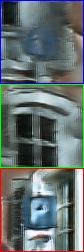}
	\endminipage\hfill
	\minipage{0.3976\textwidth}
	\includegraphics[width=\linewidth]{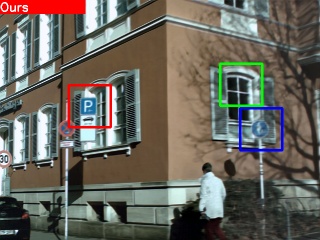}
	\endminipage\hfill
	\minipage{0.0994\textwidth}
	\includegraphics[width=\linewidth]{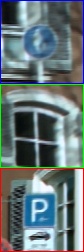}
	\endminipage\hfill
	\caption{\textbf{Qualitative comparisons on the Fastec dataset.} All the baseline networks are retrained on our Fastec dataset.}
	\label{fig_qualative_results_fastec_6}
	\vspace{0.6cm}
\end{figure*}

\begin{figure*}
	\centering
	\minipage{0.3976\textwidth}
	\includegraphics[width=\linewidth]{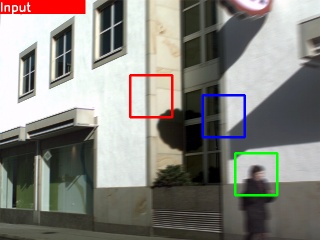}
	\endminipage\hfill
	\minipage{0.0994\textwidth}
	\includegraphics[width=\linewidth]{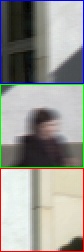}
	\endminipage\hfill
	\minipage{0.3976\textwidth}
	\includegraphics[width=\linewidth]{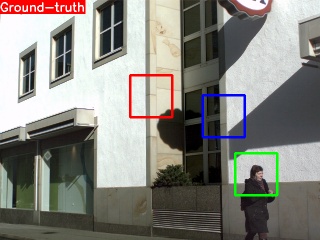}
	\endminipage\hfill
	\minipage{0.0994\textwidth}
	\includegraphics[width=\linewidth]{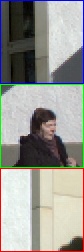}
	\endminipage\hfill
	\vspace{0.004\textheight}
	\minipage{0.3976\textwidth}
	\includegraphics[width=\linewidth]{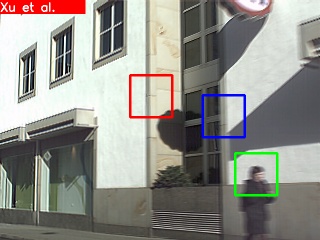}
	\endminipage\hfill
	\minipage{0.0994\textwidth}
	\includegraphics[width=\linewidth]{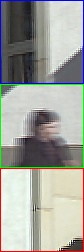}
	\endminipage\hfill
	\minipage{0.3976\textwidth}
	\includegraphics[width=\linewidth]{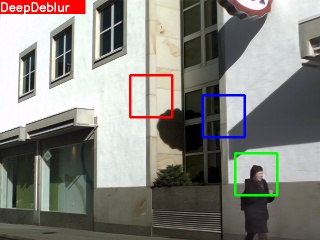}
	\endminipage\hfill
	\minipage{0.0994\textwidth}
	\includegraphics[width=\linewidth]{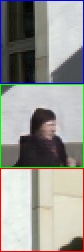}
	\endminipage\hfill
	\vspace{0.004\textheight}
	\minipage{0.3976\textwidth}
	\includegraphics[width=\linewidth]{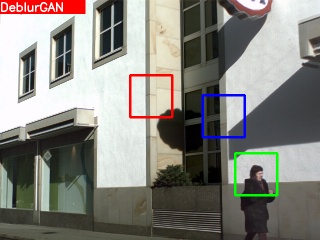}
	\endminipage\hfill
	\minipage{0.0994\textwidth}
	\includegraphics[width=\linewidth]{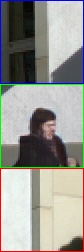}
	\endminipage\hfill
	\minipage{0.3976\textwidth}
	\includegraphics[width=\linewidth]{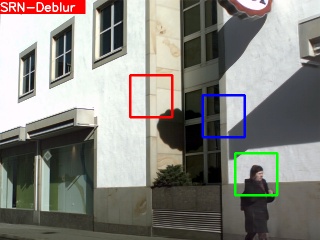}
	\endminipage\hfill
	\minipage{0.0994\textwidth}
	\includegraphics[width=\linewidth]{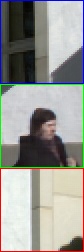}
	\endminipage\hfill
	\vspace{0.004\textheight}
	\minipage{0.3976\textwidth}
	\includegraphics[width=\linewidth]{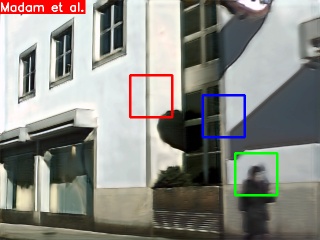}
	\endminipage\hfill
	\minipage{0.0994\textwidth}
	\includegraphics[width=\linewidth]{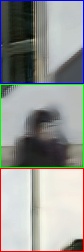}
	\endminipage\hfill
	\minipage{0.3976\textwidth}
	\includegraphics[width=\linewidth]{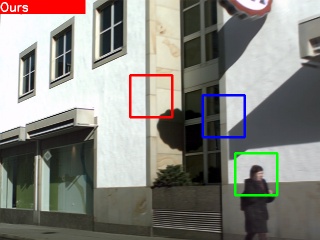}
	\endminipage\hfill
	\minipage{0.0994\textwidth}
	\includegraphics[width=\linewidth]{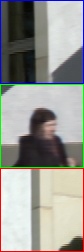}
	\endminipage\hfill
	\caption{\textbf{Qualitative comparisons on the Fastec dataset.} All the baseline networks are retrained on our Fastec dataset.}
	\label{fig_qualative_results_fastec_7}
	\vspace{0.6cm}
\end{figure*}

In \figref{fig_qualative_results_fastec_1} to \figref{fig_qualative_results_fastec_7}, 
we present additional qualitative experimental results on single image deblurring on the synthetic dataset. 
The results demonstrate that our method can generate visually compelling sharp images 
that are competitive to three state-of-the-art supervised methods \cite{Supp_Nah2017CVPR, Supp_Kupyn2018CVPR, Supp_Tao2018CVPR}. For fair comparisons, we retrain all the networks on our Fastec dataset. It also significantly outperforms the state-of-the-art optimization-based method from Xu et al. \cite{Supp_Xu2013CVPR} and the self-supervised method from \cite{Supp_Madam2018ECCV}. 
The optimization based method from Xu et al.\cite{Supp_Xu2013CVPR} fails to deblur blurry images from this dataset. To make the problem tractable, they assume that the motion blur is caused by either camera rotation or in-plane translation. However, those assumptions are violated in our Fastec dataset, where the motion blur is also caused by the 3D scene geometry.

\begin{figure*}
	\centering
	\minipage{0.33\textwidth}
	\includegraphics[width=\linewidth]{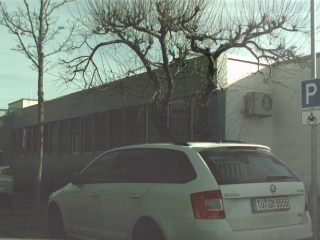}
	\endminipage\hfill
	\minipage{0.33\textwidth}
	\includegraphics[width=\linewidth]{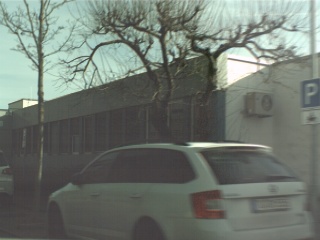}
	\endminipage\hfill
	\minipage{0.33\textwidth}
	\includegraphics[width=\linewidth]{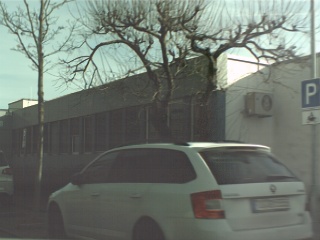}
	\endminipage\hfill
	%
	\minipage{0.33\textwidth}
	\includegraphics[width=\linewidth]{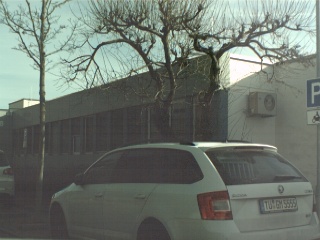}
	\endminipage\hfill
	\minipage{0.33\textwidth}
	\includegraphics[width=\linewidth]{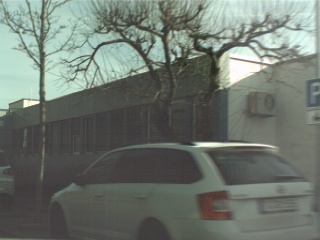}
	\endminipage\hfill
	\minipage{0.33\textwidth}
	\includegraphics[width=\linewidth]{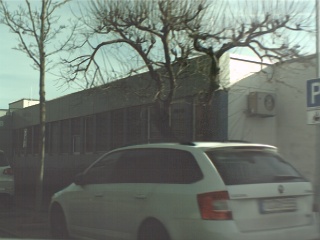}
	\endminipage\hfill
	%
	\minipage{0.33\textwidth}
	\includegraphics[width=\linewidth]{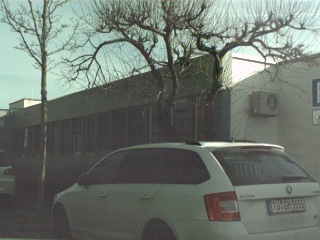}
	\endminipage\hfill
	\minipage{0.33\textwidth}
	\includegraphics[width=\linewidth]{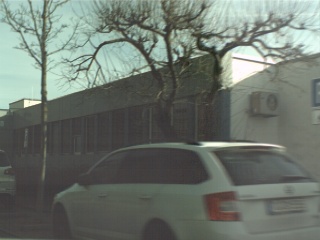}
	\endminipage\hfill
	\minipage{0.33\textwidth}
	\includegraphics[width=\linewidth]{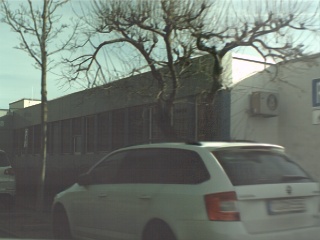}
	\endminipage\hfill
	%
	\minipage{0.33\textwidth}
	\includegraphics[width=\linewidth]{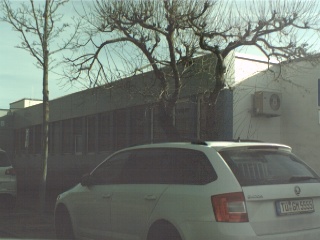}
	\endminipage\hfill
	\minipage{0.33\textwidth}
	\includegraphics[width=\linewidth]{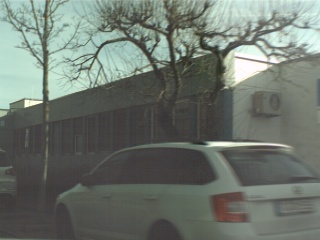}
	\endminipage\hfill
	\minipage{0.33\textwidth}
	\includegraphics[width=\linewidth]{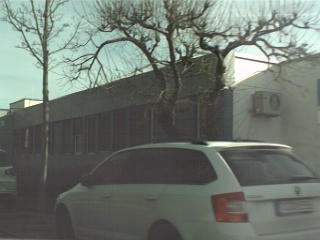}
	\endminipage\hfill
	%
	\minipage{0.33\textwidth}
	\includegraphics[width=\linewidth]{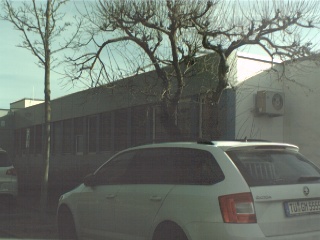}
	\endminipage\hfill
	\minipage{0.33\textwidth}
	\includegraphics[width=\linewidth]{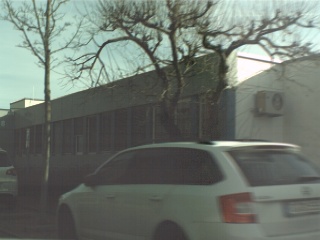}
	\endminipage\hfill
	\minipage{0.33\textwidth}
	\includegraphics[width=\linewidth]{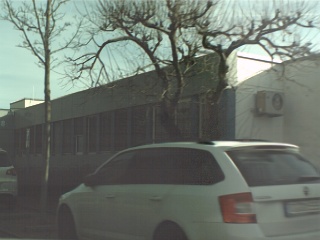}
	\endminipage\hfill
	\caption{\textbf{Temporal consistency on the Fastec dataset (frame 1-5).} \textbf{Left:} Ground truth. \textbf{Middle:} Blurry image. \textbf{Right:} Deblurred image by our network.}
	\label{fig_tc_results_fastec_1}
\end{figure*}

\begin{figure*}
	\centering
	\minipage{0.33\textwidth}
	\includegraphics[width=\linewidth]{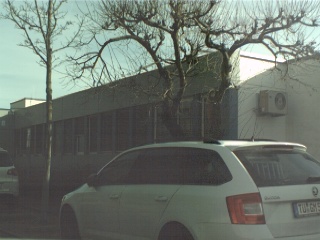}
	\endminipage\hfill
	\minipage{0.33\textwidth}
	\includegraphics[width=\linewidth]{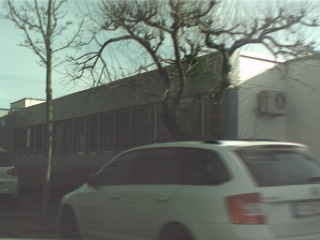}
	\endminipage\hfill
	\minipage{0.33\textwidth}
	\includegraphics[width=\linewidth]{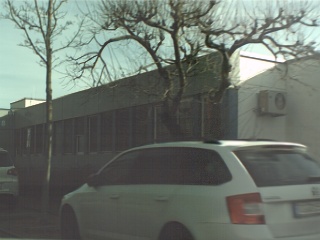}
	\endminipage\hfill
	%
	\minipage{0.33\textwidth}
	\includegraphics[width=\linewidth]{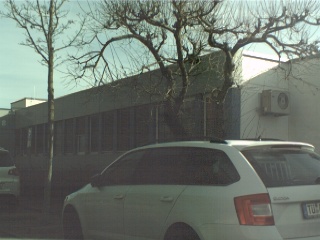}
	\endminipage\hfill
	\minipage{0.33\textwidth}
	\includegraphics[width=\linewidth]{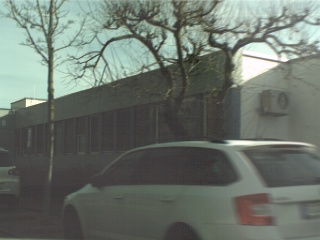}
	\endminipage\hfill
	\minipage{0.33\textwidth}
	\includegraphics[width=\linewidth]{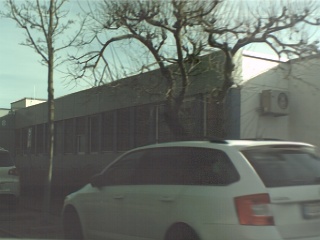}
	\endminipage\hfill
	%
	\minipage{0.33\textwidth}
	\includegraphics[width=\linewidth]{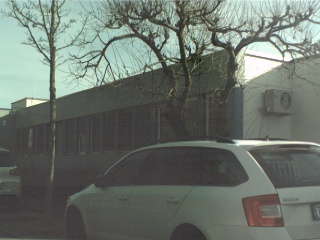}
	\endminipage\hfill
	\minipage{0.33\textwidth}
	\includegraphics[width=\linewidth]{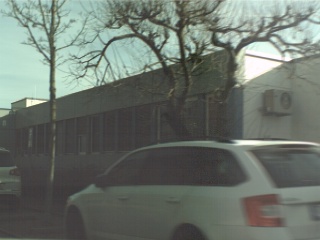}
	\endminipage\hfill
	\minipage{0.33\textwidth}
	\includegraphics[width=\linewidth]{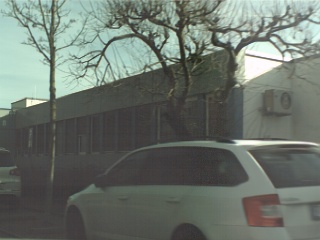}
	\endminipage\hfill
	%
	\minipage{0.33\textwidth}
	\includegraphics[width=\linewidth]{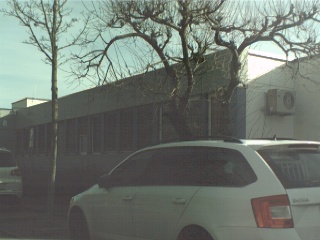}
	\endminipage\hfill
	\minipage{0.33\textwidth}
	\includegraphics[width=\linewidth]{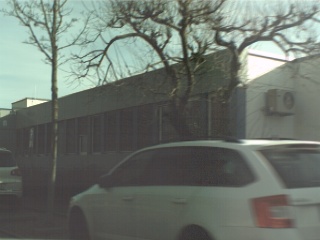}
	\endminipage\hfill
	\minipage{0.33\textwidth}
	\includegraphics[width=\linewidth]{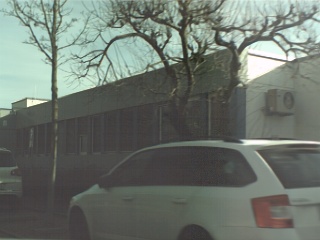}
	\endminipage\hfill
	%
	\minipage{0.33\textwidth}
	\includegraphics[width=\linewidth]{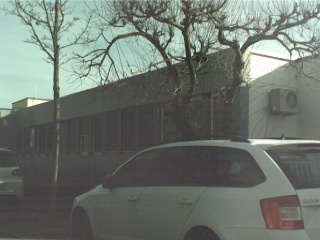}
	\endminipage\hfill
	\minipage{0.33\textwidth}
	\includegraphics[width=\linewidth]{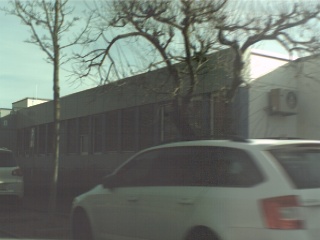}
	\endminipage\hfill
	\minipage{0.33\textwidth}
	\includegraphics[width=\linewidth]{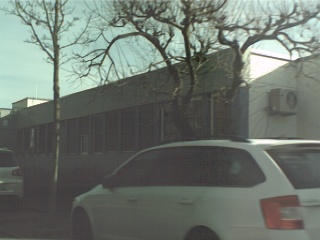}
	\endminipage\hfill
	\caption{\textbf{Temporal consistency on the Fastec dataset (frame 6-10).} \textbf{Left:} Ground truth. \textbf{Middle:} Blurry image. \textbf{Right:} Deblurred image by our network.}
	\label{fig_tc_results_fastec_2}
\end{figure*}

\begin{figure*}
	\centering
	\minipage{0.33\textwidth}
	\includegraphics[width=\linewidth]{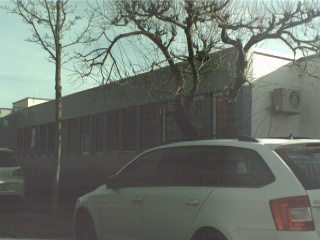}
	\endminipage\hfill
	\minipage{0.33\textwidth}
	\includegraphics[width=\linewidth]{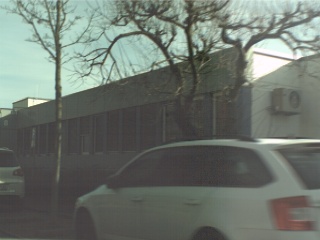}
	\endminipage\hfill
	\minipage{0.33\textwidth}
	\includegraphics[width=\linewidth]{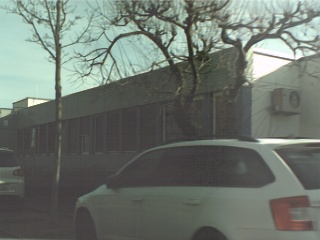}
	\endminipage\hfill
	%
	\minipage{0.33\textwidth}
	\includegraphics[width=\linewidth]{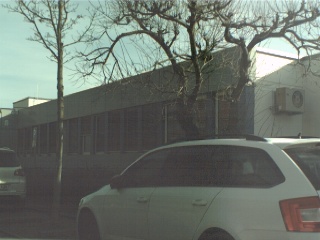}
	\endminipage\hfill
	\minipage{0.33\textwidth}
	\includegraphics[width=\linewidth]{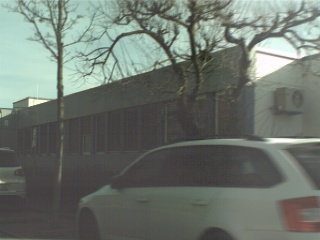}
	\endminipage\hfill
	\minipage{0.33\textwidth}
	\includegraphics[width=\linewidth]{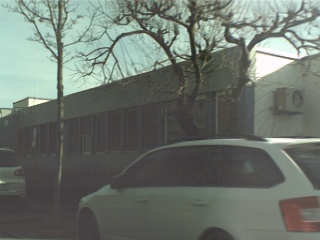}
	\endminipage\hfill
	%
	\minipage{0.33\textwidth}
	\includegraphics[width=\linewidth]{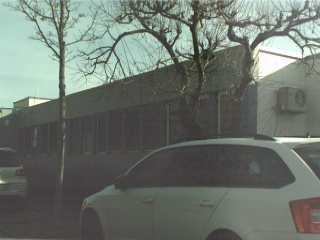}
	\endminipage\hfill
	\minipage{0.33\textwidth}
	\includegraphics[width=\linewidth]{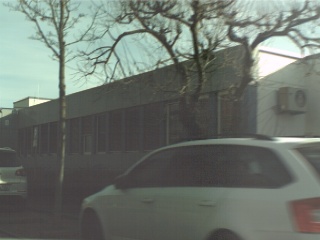}
	\endminipage\hfill
	\minipage{0.33\textwidth}
	\includegraphics[width=\linewidth]{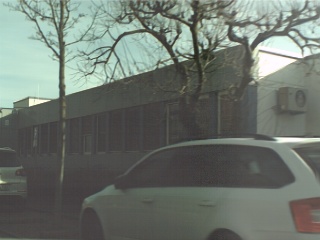}
	\endminipage\hfill
	%
	\minipage{0.33\textwidth}
	\includegraphics[width=\linewidth]{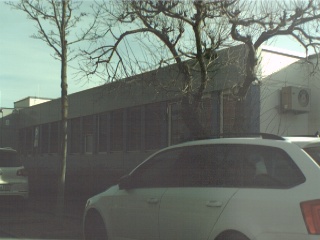}
	\endminipage\hfill
	\minipage{0.33\textwidth}
	\includegraphics[width=\linewidth]{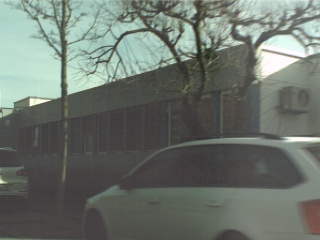}
	\endminipage\hfill
	\minipage{0.33\textwidth}
	\includegraphics[width=\linewidth]{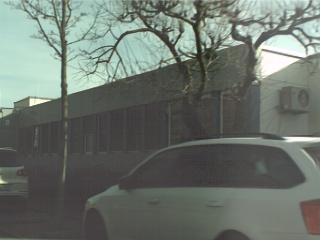}
	\endminipage\hfill
	%
	\minipage{0.33\textwidth}
	\includegraphics[width=\linewidth]{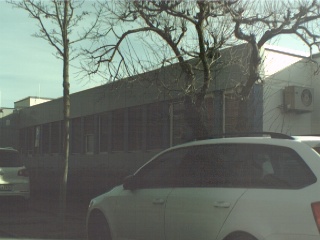}
	\endminipage\hfill
	\minipage{0.33\textwidth}
	\includegraphics[width=\linewidth]{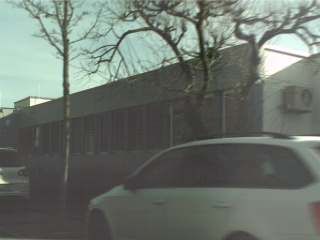}
	\endminipage\hfill
	\minipage{0.33\textwidth}
	\includegraphics[width=\linewidth]{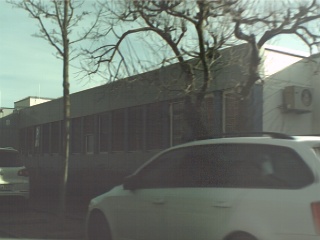}
	\endminipage\hfill
	\caption{\textbf{Temporal consistency on the Fastec dataset (frame 11-15).} \textbf{Left:} Ground truth. \textbf{Middle:} Blurry image. \textbf{Right:} Deblurred image by our network.}
	\label{fig_tc_results_fastec_3}
\end{figure*}

\begin{figure*}
	\centering
	\minipage{0.33\textwidth}
	\includegraphics[width=\linewidth]{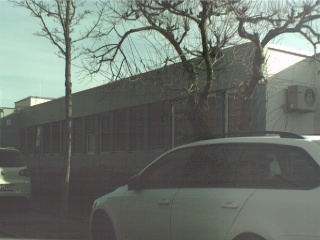}
	\endminipage\hfill
	\minipage{0.33\textwidth}
	\includegraphics[width=\linewidth]{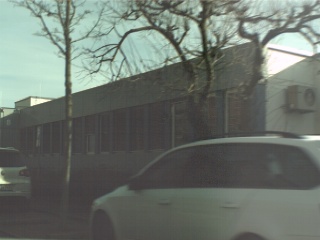}
	\endminipage\hfill
	\minipage{0.33\textwidth}
	\includegraphics[width=\linewidth]{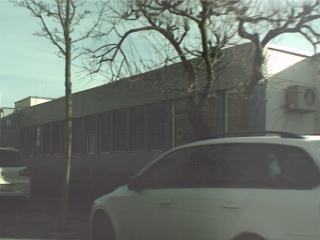}
	\endminipage\hfill
	%
	\minipage{0.33\textwidth}
	\includegraphics[width=\linewidth]{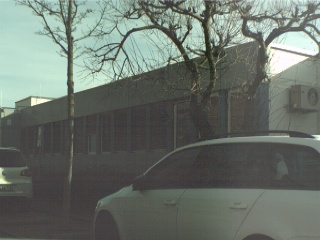}
	\endminipage\hfill
	\minipage{0.33\textwidth}
	\includegraphics[width=\linewidth]{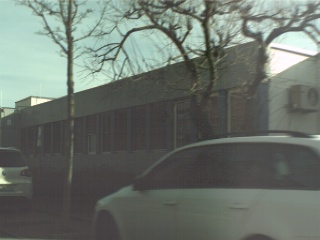}
	\endminipage\hfill
	\minipage{0.33\textwidth}
	\includegraphics[width=\linewidth]{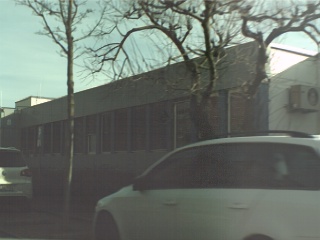}
	\endminipage\hfill
	%
	\minipage{0.33\textwidth}
	\includegraphics[width=\linewidth]{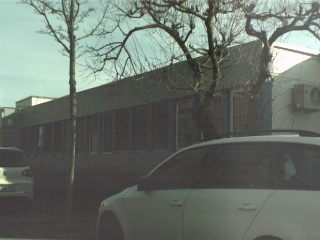}
	\endminipage\hfill
	\minipage{0.33\textwidth}
	\includegraphics[width=\linewidth]{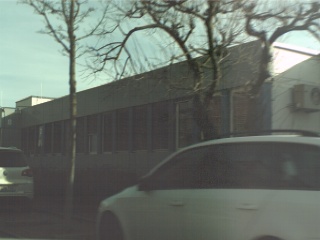}
	\endminipage\hfill
	\minipage{0.33\textwidth}
	\includegraphics[width=\linewidth]{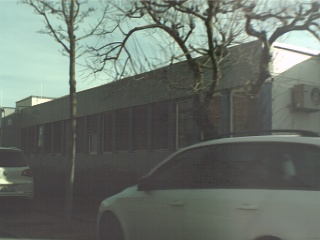}
	\endminipage\hfill
	%
	\minipage{0.33\textwidth}
	\includegraphics[width=\linewidth]{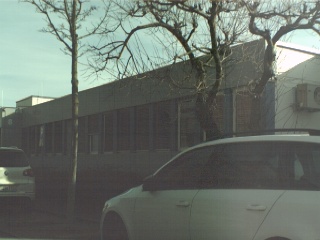}
	\endminipage\hfill
	\minipage{0.33\textwidth}
	\includegraphics[width=\linewidth]{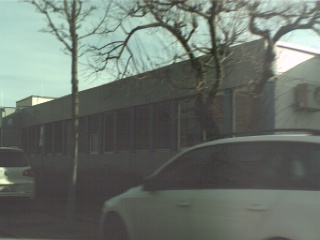}
	\endminipage\hfill
	\minipage{0.33\textwidth}
	\includegraphics[width=\linewidth]{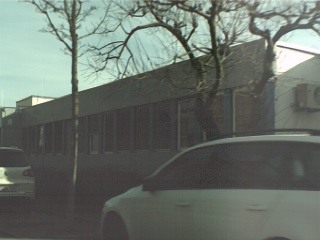}
	\endminipage\hfill
	\caption{\textbf{Temporal consistency on the Fastec dataset (frame 16-19).} \textbf{Left:} Ground truth. \textbf{Middle:} Blurry image. \textbf{Right:} Deblurred image by our network.}
	\label{fig_tc_results_fastec_4}
\end{figure*}

\begin{figure*}
	\centering
	\minipage{0.33\textwidth}
	\includegraphics[width=\linewidth]{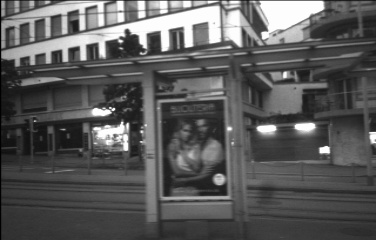}
	\endminipage\hfill
	\minipage{0.33\textwidth}
	\includegraphics[width=\linewidth]{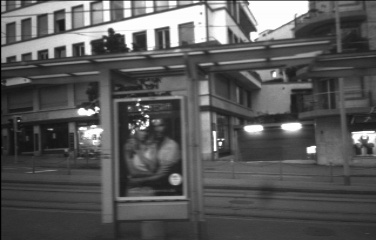}
	\endminipage\hfill
	\minipage{0.33\textwidth}
	\includegraphics[width=\linewidth]{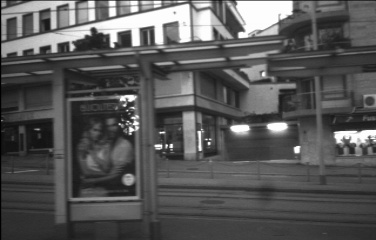}
	\endminipage\hfill
	\minipage{0.33\textwidth}
	\includegraphics[width=\linewidth]{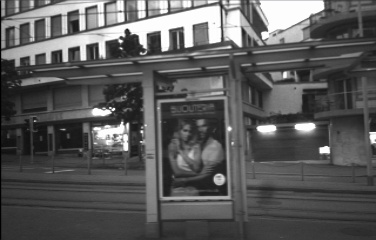}
	\endminipage\hfill
	\minipage{0.33\textwidth}
	\includegraphics[width=\linewidth]{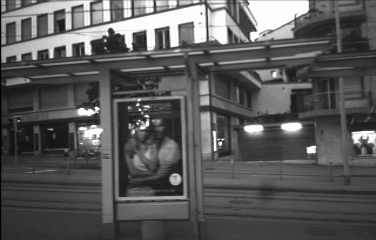}
	\endminipage\hfill
	\minipage{0.33\textwidth}
	\includegraphics[width=\linewidth]{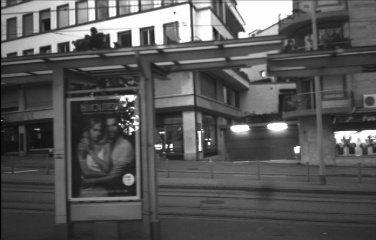}
	\endminipage\hfill
	\minipage{0.33\textwidth}
	\includegraphics[width=\linewidth]{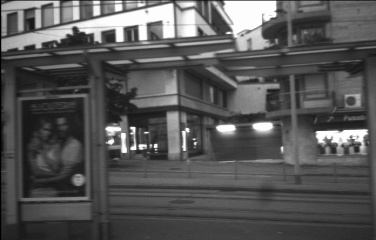}
	\endminipage\hfill
	\minipage{0.33\textwidth}
	\includegraphics[width=\linewidth]{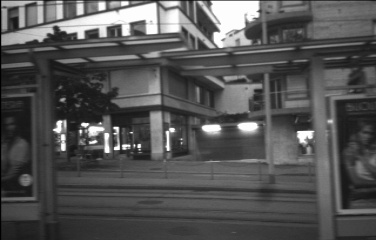}
	\endminipage\hfill
	\minipage{0.33\textwidth}
	\includegraphics[width=\linewidth]{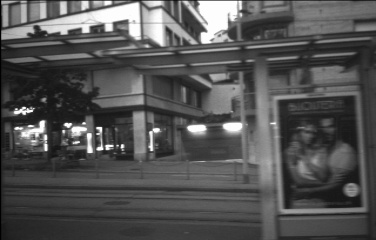}
	\endminipage\hfill
	\minipage{0.33\textwidth}
	\includegraphics[width=\linewidth]{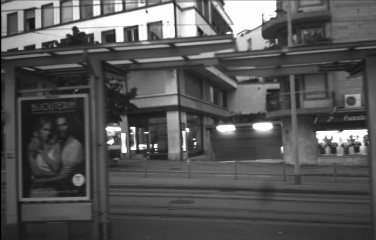}
	\endminipage\hfill
	\minipage{0.33\textwidth}
	\includegraphics[width=\linewidth]{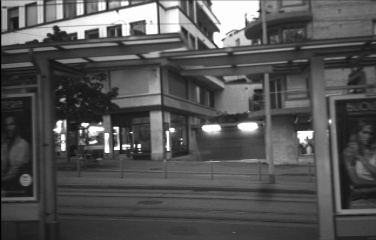}
	\endminipage\hfill
	\minipage{0.33\textwidth}
	\includegraphics[width=\linewidth]{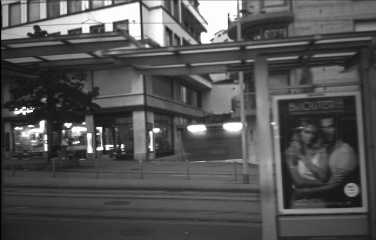}
	\endminipage\hfill
	%
	\minipage{0.33\textwidth}
	\includegraphics[width=\linewidth]{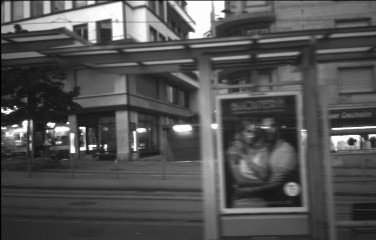}
	\endminipage\hfill
	\minipage{0.33\textwidth}
	\includegraphics[width=\linewidth]{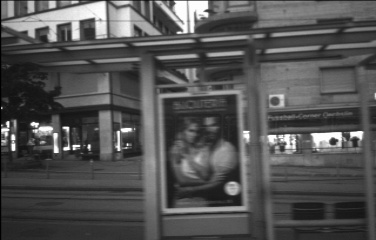}
	\endminipage\hfill
	\minipage{0.33\textwidth}
	\includegraphics[width=\linewidth]{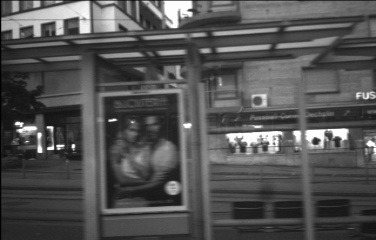}
	\endminipage\hfill
	\minipage{0.33\textwidth}
	\includegraphics[width=\linewidth]{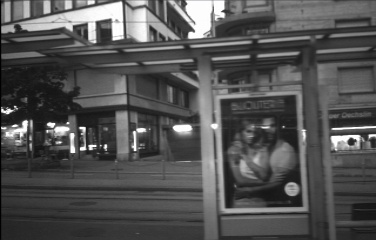}
	\endminipage\hfill
	\minipage{0.33\textwidth}
	\includegraphics[width=\linewidth]{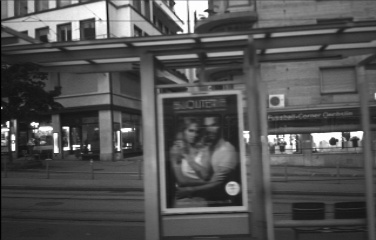}
	\endminipage\hfill
	\minipage{0.33\textwidth}
	\includegraphics[width=\linewidth]{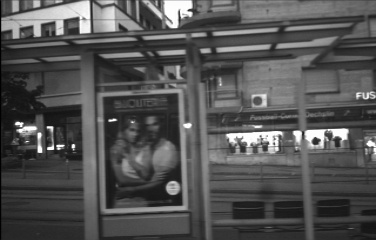}
	\endminipage\hfill
	\caption{\textbf{Temporal consistency on the real dataset.} \textbf{Odd rows:} Blurry image. \textbf{Even rows:} Deblurred image by our network.}
	\label{fig_tc_results_real_1}
\end{figure*}

\begin{figure*}
	\centering
	\minipage{0.33\textwidth}
	\includegraphics[width=\linewidth]{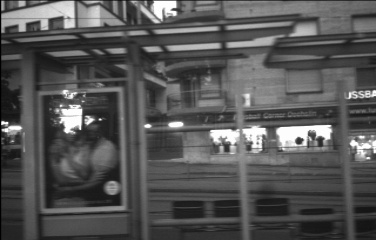}
	\endminipage\hfill
	\minipage{0.33\textwidth}
	\includegraphics[width=\linewidth]{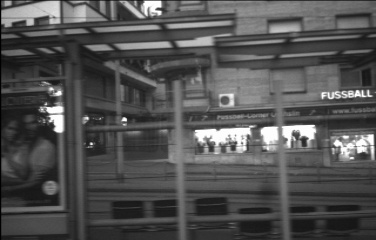}
	\endminipage\hfill
	\minipage{0.33\textwidth}
	\includegraphics[width=\linewidth]{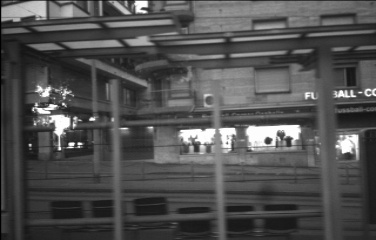}
	\endminipage\hfill
	\minipage{0.33\textwidth}
	\includegraphics[width=\linewidth]{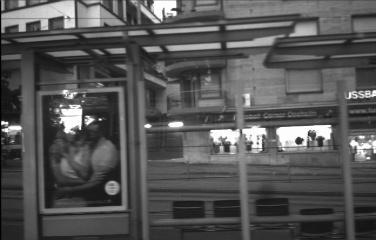}
	\endminipage\hfill
	\minipage{0.33\textwidth}
	\includegraphics[width=\linewidth]{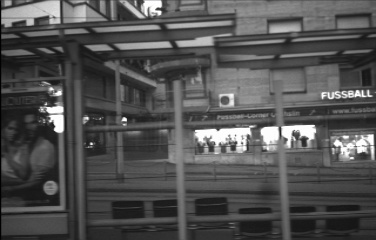}
	\endminipage\hfill
	\minipage{0.33\textwidth}
	\includegraphics[width=\linewidth]{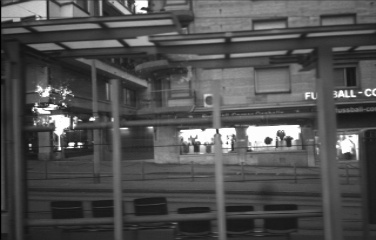}
	\endminipage\hfill
	\minipage{0.33\textwidth}
	\includegraphics[width=\linewidth]{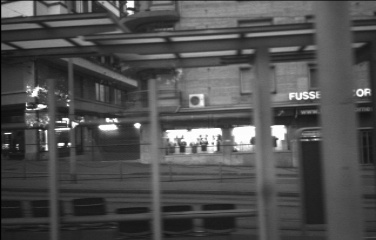}
	\endminipage\hfill
	\minipage{0.33\textwidth}
	\includegraphics[width=\linewidth]{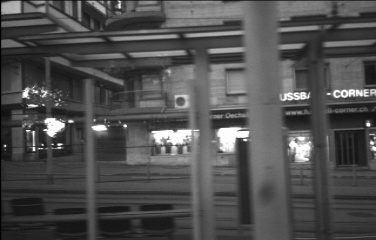}
	\endminipage\hfill
	\minipage{0.33\textwidth}
	\includegraphics[width=\linewidth]{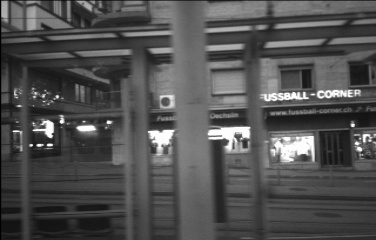}
	\endminipage\hfill
	\minipage{0.33\textwidth}
	\includegraphics[width=\linewidth]{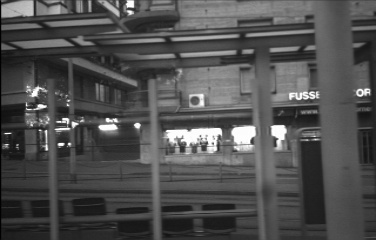}
	\endminipage\hfill
	\minipage{0.33\textwidth}
	\includegraphics[width=\linewidth]{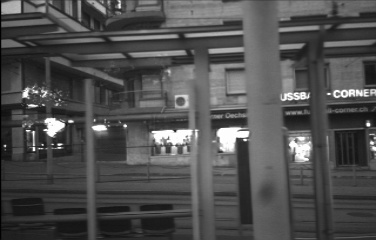}
	\endminipage\hfill
	\minipage{0.33\textwidth}
	\includegraphics[width=\linewidth]{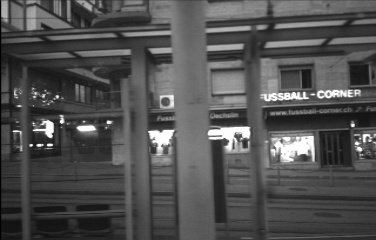}
	\endminipage\hfill
	%
	\minipage{0.33\textwidth}
	\includegraphics[width=\linewidth]{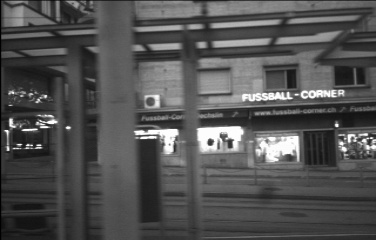}
	\endminipage\hfill
	\minipage{0.33\textwidth}
	\includegraphics[width=\linewidth]{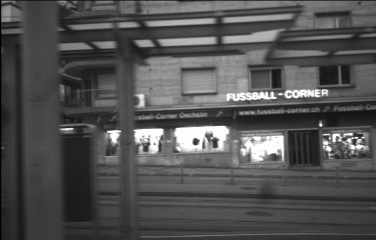}
	\endminipage\hfill
	\minipage{0.33\textwidth}
	\includegraphics[width=\linewidth]{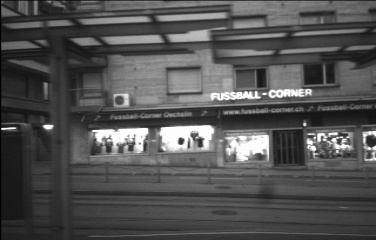}
	\endminipage\hfill
	\minipage{0.33\textwidth}
	\includegraphics[width=\linewidth]{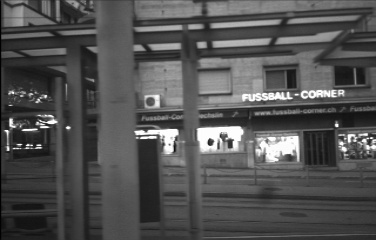}
	\endminipage\hfill
	\minipage{0.33\textwidth}
	\includegraphics[width=\linewidth]{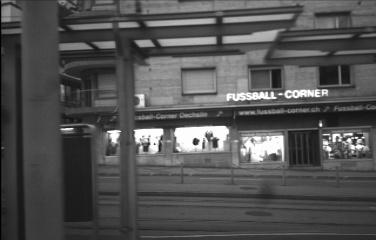}
	\endminipage\hfill
	\minipage{0.33\textwidth}
	\includegraphics[width=\linewidth]{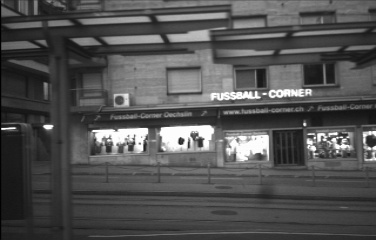}
	\endminipage\hfill
	\caption{\textbf{Temporal consistency on the real dataset.} \textbf{Odd rows:} Blurry image. \textbf{Even rows:} Deblurred image by our network.}
	\label{fig_tc_results_real_2}
\end{figure*}

To further demonstrate the temporal consistency of our method, we also present the experimental results for image sequences from \figref{fig_tc_results_fastec_1} to \figref{fig_tc_results_real_2}. The experimental results demonstrate that our network can deblur an image sequence temporally consistent, on both the synthetic and real datasets.

%

\end{document}